\documentclass[letterpaper, 10 pt, journal, twoside]{IEEEtran}  

\IEEEoverridecommandlockouts                              

\usepackage{siunitx}
\usepackage[dvipsnames]{xcolor}
\definecolor{OIblack}{RGB}{0, 0, 0}
\definecolor{OIgreen}{RGB}{0, 158, 115}
\definecolor{OIblue}{RGB}{0, 114, 178}
\definecolor{OIlightblue}{RGB}{86, 180, 233}
\definecolor{OIyellow}{RGB}{240, 228, 66}
\definecolor{OIorange}{RGB}{230, 159, 0}
\definecolor{OIred}{RGB}{213, 94, 0}
\definecolor{OIpink}{RGB}{204, 121, 167}

\colorlet{BESO}{OIgreen}
\colorlet{MPD}{OIblue}
\colorlet{DPT}{OIorange}
\colorlet{DPC}{OIpink}

\usepackage{amsmath, amssymb, latexsym}
\usepackage{bm} 

\usepackage{calc}

\usepackage{cite}

\usepackage{subcaption}
\usepackage{booktabs}

\usepackage[acronym]{glossaries}
\newacronym{mpd}{MPD}{Movement Primitive Diffusion}

\newacronym{rl}{RL}{Reinforcement Learning}
\newacronym{il}{IL}{Imitation Learning}
\newacronym{dil}{DIL}{Diffusion-based Imitation Learning}

\newacronym{mpc}{MPC}{Model Predictive Control}
\newacronym[firstplural={Degrees of Freedom} (DoF)]{dof}{DoF}{Degree of Freedom}

\newacronym{btm}{BTM}{Bimanual Tissue Manipulation}
\newacronym{glt}{GLT}{Grasp Lift Touch}
\newacronym{ll}{LL}{Ligating Loop}
\newacronym{rt}{RT}{Rope Threading}

\newacronym{ras}{RAS}{Robot-Assisted Surgery}

\newacronym{mp}{MP}{Movement Primitive}
\newacronym{dmp}{DMP}{Dynamic Movement Primitive}
\newacronym{promp}{ProMP}{Probabilistic Movement Primitive}
\newacronym{prodmp}{ProDMP}{Probabilistic Dynamic Movement Primitive}

\newacronym{ddpm}{DDPM}{Denoising Diffusion Probabilistic Model}
\newacronym{sdm}{SDM}{Score-Based Diffusion Model}
\newacronym{sgm}{SGM}{Score-Based Generative Model}

\newacronym{ode}{ODE}{Ordinary Differential Equation}
\newacronym{sde}{SDE}{Stochastic Differential Equation}

\usepackage{tikz}
\usetikzlibrary{backgrounds}
\usetikzlibrary{fit}
\usetikzlibrary{spy}
\usetikzlibrary{positioning} 
\usetikzlibrary{calc}
\usetikzlibrary{patterns} 
\usetikzlibrary{quotes} 
\usetikzlibrary{calc}
\usetikzlibrary{decorations.markings}
\usepackage{pgfplots}
\pgfplotsset{compat=1.16}
\usepgfplotslibrary{fillbetween}

\usetikzlibrary{fit}
\tikzset{
  fitting node/.style={
    inner sep=0pt,
    fill=none,
    draw=none,
    reset transform,
    fit={(\pgf@pathminx,\pgf@pathminy) (\pgf@pathmaxx,\pgf@pathmaxy)}
  },
  reset transform/.code={\pgftransformreset}
}

\usepackage{xspace}
\newcommand*{\eg}{\emph{e.g.}\@\xspace}

\newcommand*{\ie}{\emph{i.e.}\@\xspace}

\newcommand*\diff{\mathop{}\!\mathrm{d}}

\usepackage{hyperref}
\hypersetup{%
    colorlinks=false,
    allbordercolors={1 1 1},
    hidelinks
}

\def\equationautorefname#1#2\null{%
  Eq.#1(#2\null)%
} 

\title{Movement Primitive Diffusion: Learning Gentle Robotic Manipulation of Deformable Objects}

\author{Paul Maria Scheikl$^{1}$, Nicolas Schreiber$^{2}$, Christoph Haas$^{2}$, Niklas Freymuth$^{2}$,\\Gerhard Neumann$^{2}$, Rudolf Lioutikov$^{2}$, and Franziska Mathis-Ullrich$^{1}$%
\thanks{Manuscript received: December 12, 2023; Revised February 23, 2024; Accepted March 18, 2024.
This paper was recommended for publication by Editor Pietro Valdastri upon evaluation of the Associate Editor and Reviewers' comments.
This work was supported by the Erlangen National High Performance Computing Center funded by the German Research Foundation (DFG), the HoreKa supercomputer funded by the Ministry of Science, Research and the Arts Baden-Württemberg and by the Federal Ministry of Education and Research, and the DFG – 448648559.}%
\thanks{$^{1}$ P. M. Scheikl and F. Mathis-Ullrich are with the Department Artificial Intelligence in Biomedical Engineering, Friedrich-Alexander-University Erlangen-Nürnberg, 91054 Erlangen, Germany. \newline {\small Corresponding author: \tt franziska.mathis-ullrich@fau.de}}%
\thanks{$^{2}$ N. Schreiber, C. Haas, N. Freymuth, G. Neumann, and R. Lioutikov are with the Institute for Anthropomatics and Robotics, Karlsruhe Institute of Technology, 76131 Karlsruhe, Germany.}%
\thanks{Digital Object Identifier (DOI): see top of this page.}
}

\begin{document}

\markboth{IEEE Robotics and Automation Letters. Preprint Version. Accepted March, 2024}
{Scheikl \MakeLowercase{\textit{et al.}}: Movement Primitive Diffusion} 

\maketitle

\begin{abstract}
    Policy learning in robot-assisted surgery (RAS) lacks data efficient and versatile methods that exhibit the desired motion quality for delicate surgical interventions.
    To this end, we introduce Movement Primitive Diffusion (MPD), a novel method for imitation learning (IL) in RAS that focuses on gentle manipulation of deformable objects.
    The approach combines the versatility of diffusion-based imitation learning (DIL) with the high-quality motion generation capabilities of Probabilistic Dynamic Movement Primitives (ProDMPs).
    This combination enables MPD to achieve gentle manipulation of deformable objects, while maintaining data efficiency critical for RAS applications where demonstration data is scarce.
    We evaluate MPD across various simulated and real world robotic tasks on both state and image observations.
    MPD outperforms state-of-the-art DIL methods in success rate, motion quality, and data efficiency.\\
    Project page: \href{https://scheiklp.github.io/movement-primitive-diffusion/}{scheiklp.github.io/movement-primitive-diffusion}
\end{abstract}

\section{INTRODUCTION}
\label{sec:introduction}
    \IEEEPARstart{A}{dvancing} the level of autonomy in \gls{ras} requires novel methods for training policies that satisfy the special requirements of surgical applications.
    \gls{ras} requires the policies to exhibit gentle manipulation of delicate tissue and perform with limited data as human demonstrations are costly.
    Additionally, human behavior is inherently multimodal~\cite{blessing2023information}, covering multiple distinct strategies for solving the same task.
    \gls{il} methods that are unable to represent multimodal behavior may exhibit harmful behavior through mode averaging that is unacceptable in surgical settings, \eg, by averaging over two distinct strategies of dissecting tissue and thus damaging healthy tissue.
    \Gls{dil} has shown to perform well on high-dimensional action spaces, generate multimodal behaviors, and exhibit strong training stability~\cite{chi2023diffusionpolicy, reuss2023multimodal}, making it a promising framework for application in \gls{ras}.
    
    \Gls{dil} methods train large neural networks to iteratively denoise action sequences drawn from a prior Gaussian distribution to generate motion conditioned on observations.
    We propose to add temporal correlations between actions during motion generation by utilizing \glspl{mp} to address both gentle manipulation of deformable objects and data efficiency in \gls{dil}.
    In other methods, neural networks output actions sequences directly~\cite{chi2023diffusionpolicy, reuss2023multimodal}.
    In our proposed method, \gls{mpd}, the neural network outputs parameters of a \gls{mp} that encode a denoised action sequence.
    These parameters are decoded into smooth position trajectories to enable gentle manipulation of deformable objects.
    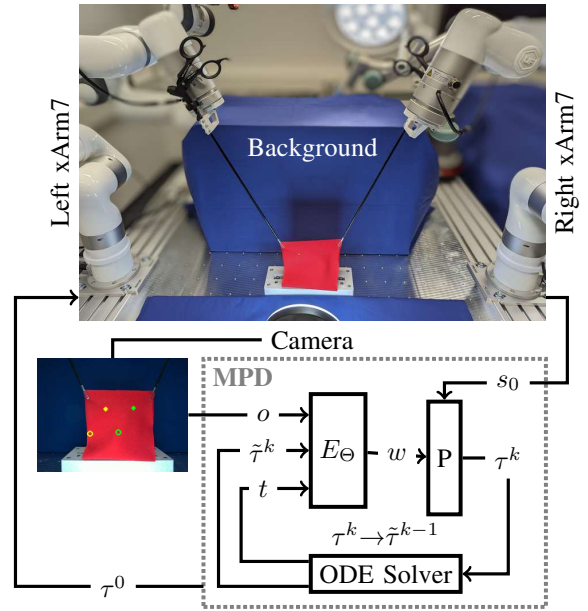
\begin{figure}
    \newdimen\EncoderWestX
    \newdimen\TauWestY
    \newdimen\TWestY
    \newdimen\ObsWestY
    \newdimen\PNorthX
    \newdimen\ODEEastY
    \newdimen\RWNorthY
    \newdimen\RWLeftY
    \newdimen\BoxWestX
    \newdimen\RGBNorthX
    \newdimen\LRobotNorthY
\centering
\vspace{1.5mm}
\begin{tikzpicture}

    \tikzset{every edge quotes/.style =
        { fill = white,
        execute at begin node = $,
        execute at end node   = $  }}
            
    \begin{scope}[local bounding box=mpd scope]
        \node (obs) {$o$};
        \pgfextracty{\ObsWestY}{\pgfpointanchor{obs}{west}}
        \node[below = 0cm of obs.south, anchor=north] (tau) {$\tilde{\tau}^k$};
        \pgfextracty{\TauWestY}{\pgfpointanchor{tau}{west}}
        \node[below = 0cm of tau.south, anchor=north] (t) {$t$};
        \pgfextracty{\TWestY}{\pgfpointanchor{t}{west}}
        
        \node[line width=1.5pt, draw, right = 0.3cm of tau.east, minimum height=1.5cm] (E) {$E_\Theta$};
        \pgfextractx{\EncoderWestX}{\pgfpointanchor{E}{west}}

        \draw [line width=1.2pt, ->] (tau.east) -- (\EncoderWestX, \the\tikz@lastysaved);
        \draw [line width=1.2pt,->] (obs.east) -- (\EncoderWestX, \the\tikz@lastysaved);
        \draw [line width=1.2pt,->] (t.east) -- (\EncoderWestX, \the\tikz@lastysaved);

        \node[line width=1.5pt, draw, below = 0.6cm of E.south west, anchor=north west] (ODE) {ODE Solver};
        \pgfextracty{\ODEEastY}{\pgfpointanchor{ODE}{east}}
        \node[above = 0.0cm of ODE.north] {$\tau^k$\textrightarrow$\tilde{\tau}^{k-1}$};

        \node[line width=1.5pt, draw, above = 0.5cm of ODE.north east, anchor=south east, minimum height=1.5cm] (P) {P};
        \pgfextractx{\PNorthX}{\pgfpointanchor{P}{north}}
        \draw[line width=1.2pt,->] (E) to ["w"] (P); 

        \draw[line width=1.2pt] ([yshift=-0.1cm, xshift=0pt]ODE.north west) --  ++(-0.9cm, 0.0) -- (\the\tikz@lastxsaved, \TWestY) -- (t.west);
        \draw[line width=1.2pt] ([yshift=0.1cm, xshift=0pt]ODE.south west) --  ++(-1.2cm, 0.0) -- (\the\tikz@lastxsaved, \TauWestY) -- (tau.west);

        \node[right = 0.3cm of P.east] (denoised) {$\tau^k$};
        \draw[line width=1.2pt] (P.east) -- (denoised.west);
        \draw[line width=1.2pt, ->] (denoised.south) -- (\the\tikz@lastxsaved, \ODEEastY) -- (ODE.east);
        
        \node[above = 0.5cm of denoised.north, anchor=south] (s) {$s_0$};
        \draw[line width=1.2pt, ->] (s.west) -- (\PNorthX, \the\tikz@lastysaved) -- (P.north);
        
    \end{scope}

    \begin{scope}[local bounding box=box scope]
        \draw[line width=1.5pt, gray, densely dotted] ($(mpd scope.north west)+(-0.2cm, 0.1cm)$) rectangle ($(mpd scope.south east)+(0.2cm, -0.2cm)$);
    \end{scope}
    \pgfextractx{\BoxWestX}{\pgfpointanchor{box scope}{west}}
    \node[gray, anchor=north west] at (box scope.north west) {\textbf{MPD}};

    \node[inner sep=0pt, above = 0.5 cm of box scope.north east, anchor=south east] (RW) {\includegraphics[width=0.7\columnwidth]{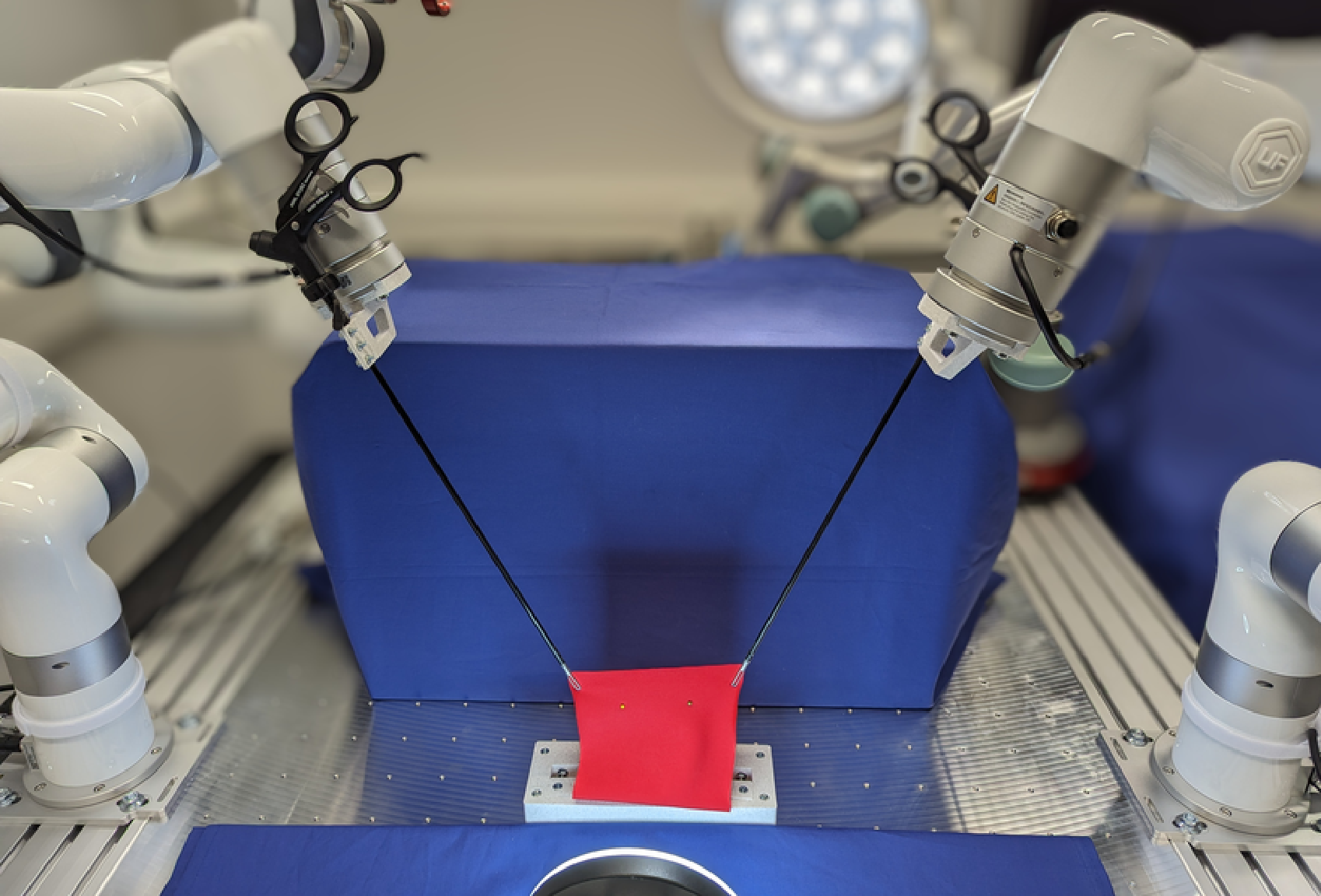}};
    \pgfextracty{\RWNorthY}{\pgfpointanchor{RW}{north}}
    \pgfextracty{\RWLeftY}{\pgfpointanchor{RW}{210}}
    
    \node[anchor=north] (cam) at (RW.south) {Camera};
    \node[anchor=south, rotate=90] (Lrobot) at (RW.west) {Left xArm7};
    \pgfextracty{\LRobotNorthY}{\pgfpointanchor{Lrobot}{north}}
    \node[anchor=north, rotate=90] (Rrobot) at (RW.east) {Right xArm7};
    \node[above = 0.00cm of RW.center, align=center, white, anchor=south, inner sep=0pt] {Background};

    \node[inner sep=0pt, anchor=east] (rgb obs) at ($(\BoxWestX, \ObsWestY) + (-0.2cm, 0.0cm)$) {\includegraphics[width=2.0cm]{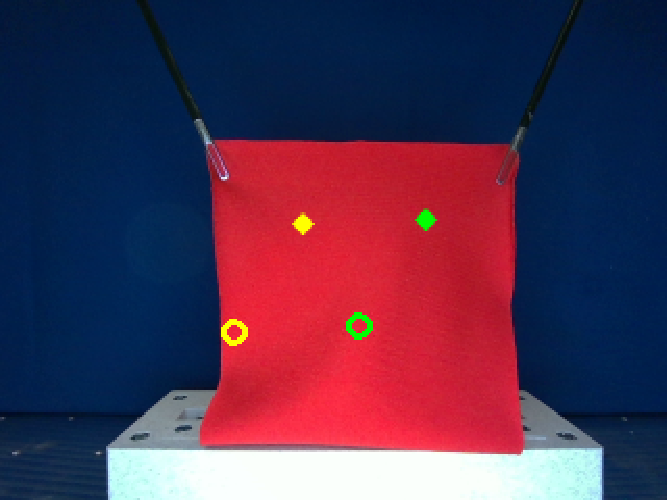}};
    \pgfextractx{\RGBNorthX}{\pgfpointanchor{rgb obs}{north}}

    \draw[line width=1.2pt] (cam.west) -| (rgb obs.north);
    \draw[line width=1.2pt] (rgb obs.east) -- (obs.west);

    \draw[line width=1.2pt] (s.east) -- ++(0.5cm, 0.0) -- (\the\tikz@lastxsaved, \RWLeftY) -- (RW.330);
    
    \node[anchor=east] (final tau) at ([yshift=0.1cm, xshift=-2.3cm]ODE.south west) {$\tau^0$};
    \draw[line width=1.2pt] (final tau.east) -- (\BoxWestX, \the\tikz@lastysaved);
    \draw[line width=1.2pt, ->] (final tau.west) -- ++(-1.0cm, 0.0) -- (\the\tikz@lastxsaved, \RWLeftY) -- (RW.210);

\end{tikzpicture}
\caption{
    Schematic for action sequence generation with MPD for bimanual tissue manipulation.
    Observations $o$ and initial values $s_0$ for position and velocity are captured on the bimanual robotic setup.
    An ODE solver solves the Probability Flow ODE with learnable model $E_\Theta$ and ProDMP $P$ by iteratively denoising an action sequence $\tilde{\tau}^k$ for diffusion step $k$ and respective noise level $t$.
    The final denoised action sequence $\tau^0$ is executed on the robots.
}
\label{fig:overview}
\end{figure}

    Leveraging both \glspl{mp} and \gls{dil}, \gls{mpd} provides a novel approach with increased data efficiency and generation of action sequences that are suitable for gentle deformable object manipulation in \gls{ras}.
    \Gls{mpd}, generates gentle motions that outperform state-of-the art \gls{dil} methods in terms of success rate, motion quality, and required training data.
    \gls{mpd} enables generating action sequences with guaranteed initial conditions for position and velocity, and benefits from a modern diffusion framework for fast inference times.
    \gls{mpd} works well for both state observations, as well as raw RGB image observations, which is a critical aspect for application in \gls{ras}, as image observations are the only readily available source of information~\cite{scheikl2023sim2real}.
    An overview of action sequence generation with \gls{mpd} is illustrated in \autoref{fig:overview}.

\section{RELATED WORK}
\label{sec:related_work}
    Deformable object manipulation in surgery is explored through either modeling deformations or predicting action sequences.
    Learning deformation models typically offers better generalization by capturing object properties rather than task-specific movements~\cite{thach2022shapecontrol}.
    However, these models commonly only learn the deformation behavior and not the behavior that is required to solve a task~\cite{shin2019autonomous}.
    They are thus not applicable to tasks that require complex motions that do not directly deform the object, for example grasping.
    Model-based strategies are effective for known goal configurations of deformable objects~\cite{thach2022shapecontrol} or when deformation involves visual servoing of identifiable landmarks~\cite{shin2019autonomous}.
    However, in actual surgical scenarios, where goal configurations and landmarks are uncertain and grasping deformable objects is crucial, these approaches may not suffice. 
    Consequently, our research prioritizes action sequence prediction to address diverse surgical tasks involving grasping, manipulation of deformable objects, and precise motion skills.

    Tissue manipulation is a common surgical task that encompasses both direct and indirect manipulation of deformable objects.
    Examples for direct manipulation are tissue retraction to visualize occluded structures~\cite{scheikl2021cooperative, pore2021learning, scheikl2023sim2real} or organ manipulation to bring a deformable object into a desired shape~\cite{thach2022shapecontrol, dettorre2022organmanipulation}.
    Indirect manipulation involves moving specific landmarks on tissue by altering the tissue's shape through manipulation of other points on the tissue.
    Shin~et~al.~\cite{shin2019autonomous} investigate \gls{il} and \gls{rl} for tissue manipulation to estimate the dynamics of an adaptive \gls{mpc}.
    They argue that \gls{il} works well if the dataset is sufficiently large and that \gls{rl} is not suited for application on a real robotic system as the learning procedure is too hazardous.
    Ou~et~al.~\cite{ou2023sim2real} address this concern through sim-to-real transfer, by randomizing parameters of the simulation to bridge the physical sim-to-real gap.
    In our own prior work~\cite{scheikl2023sim2real}, we employ a learned image translation model to bridge the visual sim-to-real gap for \gls{rl} of image-based policies for tissue retraction, but require an accurate digital twin of the real setup.
    In summary, recent work for deformable object manipulation in \gls{ras} is divided between \gls{rl} and \gls{il}.
    \gls{il} is the preferred choice for image-based policies, as modeling a digital twin and the resulting sim-to-real gap are still significant hurdles for \gls{rl} methods.
    However, the current methods are unable to learn multimodal behaviors and are not sufficiently data efficient.
    Furthermore, current methods that feature end-to-end learning from observations to end-effector movements do not include strategies to generate gentle motions.
    \Gls{mpd} is able to address these challenges, learning multimodal behavior from limited data directly from image observations, generating action sequences for gentle manipulation.
    
    Diffusion Policy~\cite{chi2023diffusionpolicy} and BESO~\cite{reuss2023beso} present the current state of the art in robotic \gls{dil}.
    Both methods iteratively denoise action sequence samples to generate motion, conditioned on observations.
    Both works evaluate their methods against multiple state-of-the-art \gls{il} methods and find that \gls{dil} methods outperform non-diffusion-based methods in terms of success rate, and excel in learning multimodal behaviors.
    In this work, we investigate Diffusion Policy and BESO under the requirements of \gls{ras} and show that \gls{mpd} addresses the shortcomings of these methods.
    For tissue dissection, another common task in \gls{ras}, Li~et~al.~\cite{li2023imitation} extend iBC~\cite{florence2021implicit} to learn trajectories as a joint distribution of image observations and instrument motions.
    However, they do not consider transitions between multiple action sequences, as the method predicts actions for short, independent dissection trajectories.
    From a motion generation perspective of a single, independent action sequence, the method is comparable to Diffusion Policy~\cite{chi2023diffusionpolicy}.
    
    Li~et~al.~\cite{li2023prodmp} show that neural networks can be utilized in \gls{il} to predict \gls{prodmp} weight vectors to generate smooth motions with guaranteed initial conditions for planning of consecutive action sequences.
    However, the method is built on maximizing the log-likelihood of the learned trajectories and is thus unable to model multimodal behaviors.
    \Gls{mpd} is able to use \glspl{prodmp} and model multimodal behaviors through \gls{dil}.

\section{METHODS}
\label{sec:methods}
    \subsection{Problem Formulation}
    \label{sec:methods:problem}
         We predict action sequences $\tau = (\tau_i)_{i=0..n}$ that consist of desired values $\tau_i \in \mathbb{R}^k$ for the next $n$ time steps relative to the current time, where $\mathbb{R}^k$ is a $k$-dimensional task space.
         Depending on the task, the task space consists of $k$ actuation \glspl{dof} such as grasper articulation, and rotations and translations of surgical instruments in relation to a remote center of motion.
         The action sequences are predicted based on observations $o = (o_j)_{j=-m+1..0}$ from the previous $m$ time steps.
         We follow an \gls{il} approach and train our models on a dataset $\mathcal{D}$ of human demonstrations $d$.
         Each demonstration is a sequence pair $(\tau_i, o_i)_{i=0..N}$ over one full task execution with $N$ time steps.
         For training, the demonstrations are split into multiple action and observation sequences of lengths $n$ and $m$, respectively.
         The demonstrations vary in length and are commonly much longer than the predicted action sequences, $n<N$.
         We focus on action sequences instead of single actions or full trajectories.
         Action sequences achieve a balance between reducing the compounding error problem of step-based approaches and the flexibility to adapt to changes that arise in real world scenarios due to variability and uncertainties.

    \subsection{Preliminaries on Movement Primitives}
    \label{sec:methods:prodmp}
        \glspl{mp} are a representation of basic elements for robotic motion that can be combined and modulated to generate complex movements~\cite{li2023prodmp}.
        They provide a framework for representing complex motor skills through simple and parametrizable models.
        \glspl{prodmp}~\cite{li2023prodmp} offer a unifying framework that overcomes the weaknesses and combines the strengths of \glspl{promp} and \glspl{dmp}.
        \glspl{prodmp} eliminate the need for costly numerical integration associated with \glspl{dmp} by utilizing precomputed position and velocity basis functions of the fundamental \gls{ode} that are valid for all trajectories.
        In \glspl{prodmp}, the positions $y$ and velocities $\dot{y}$ of a trajectory are formulated as
        \begin{equation}
            \begin{aligned}
                 y(t)&=c_1 y_1(t) + c_2 y_2(t) + \bm\Phi(t)^\top \bm w,\\
                 \dot{y}(t)&=c_1 \dot{y}_1(t) + c_2 \dot{y}_2(t) + \dot{\bm\Phi}(t)^\top \bm w   
            \end{aligned}
            \label{eq:prodmp}
        \end{equation}
        where $y_1$ and $y_2$ are the two linearly independent complementary functions of the \gls{prodmp}'s homogeneous \gls{ode}.
        $\dot{y}_1$, $\dot{y}_2$ are their respective time derivatives.
        The coefficients $c_1$ and $c_2$ are constants derived from the boundary conditions of the \gls{ode}.
        The coefficients are calculated from desired position and velocity at a specific time step, allowing for smooth transitions between action sequences, which is not possible with \glspl{promp}.
        The basis functions for position and velocity, $\bm\Phi$ and $\dot{\bm\Phi}$, are computed once and then used as fixed functions. 
        The weights $\bm w$ are a composite vector that merge the \gls{dmp}'s original weight vector with the goal attractor to which the \gls{ode} converges.
        For simplicity, \autoref{eq:prodmp} is shown for a single \gls{dof} instead for the full multi-\gls{dof} formulation.
        \glspl{prodmp} facilitate planning smooth trajectories with guaranteed boundary conditions while minimizing computational demands.

    \subsection{Preliminaries on Score-based Diffusion Models}
    \label{sec:methods:diffusion}
        The two most prominent methods in \gls{dil} follow different diffusion frameworks. 
        Diffusion Policy~\cite{chi2023diffusionpolicy} builds on the \glspl{ddpm} framework~\cite{ho2020ddpm} that focuses on reversing the diffusion process at discrete noise levels, relying on probabilistic modeling of the process as a Markov chain.
        In contrast, BESO~\cite{reuss2023multimodal} builds on the \gls{sgm} framework that describes the diffusion process as a time-continuous \gls{sde} and learns the gradient of the log probability density, \ie, the score, of the data distribution.
        In contrast to the \gls{ddpm} framework, \gls{sgm} allows for modular selection of critical components such as the noise schedule and numerical solver~\cite{karras2022elucidating} and is often computationally less expensive~\cite{reuss2023beso}. 
        Both frameworks are capable of representing multimodal distributions of action sequences~\cite{chi2023diffusionpolicy, reuss2023multimodal}.

        For \gls{mpd}, we adopt the \gls{sgm} framework and closely follow the conventions proposed in~\cite{karras2022elucidating} and~\cite{reuss2023beso}.
        The diffusion process is modelled as a Probability Flow \gls{ode}
        \begin{equation}
            \diff\tau = -\dot{\sigma}(t)\sigma(t)\nabla_{\tau} \log p(\tau | o, \sigma(t))\diff t
        \end{equation}
        in time $t$, with score function $\nabla_{\tau} \log p(\tau | o, \sigma(t))$ as the gradient of the conditional probability $p$ of action sequences $\tau$ under observations $o$ and noise schedule $\sigma(t)$.
        Here, time $t$ is the time of the Probability Flow \gls{ode} that describes the diffusion process and not the time that governs the action sequence.
        To differentiate between the two, we use the subscript $i$ to refer to time steps of an action sequence $\tau$.

        The score function is approximated as
        \begin{equation}
            \nabla_{\tau} \log p(\tau | o, \sigma(t)) \approx \frac{D_\theta(\tau, o, \sigma(t)) - \tau}{\sigma(t)^2}\text{,}
            \label{eq:denoiser}
        \end{equation}
        where $D_\theta$ is a learned denoiser function with weights $\theta$.
        Applying this approximation to the Probability Flow \gls{ode} under a commonly used linear noise schedule $\sigma(t) = t$ yields
        \begin{equation}
            \diff\tau \approx \frac{\tau - D_\theta(\tau, o, t)}{t}\diff t\text{.}
        \label{eq:approx_denoiser}
        \end{equation}

        To further increase modularity of components and enhance performance across different noise magnitude levels, the denoiser $D_\theta$ is represented as
        \begin{equation}
            D_\theta(\tau, o, t) = c_{\text{skip}}(t)\tau + c_{\text{out}}(t)F_\theta(c_{\text{in}}(t)\tau, o, c_{\text{noise}}(t))
            \label{eq:scaler_main}
        \end{equation}
        with a learnable inner model $F_\theta$ and preconditioning by noise dependent functions $c_{\text{skip}}(t)$, $c_{\text{out}}(t)$, $c_{\text{in}}(t)$ and $c_{\text{noise}}(t)$~\cite{karras2022elucidating}.
        
        The model $D_\theta$ is trained with denoising score matching~\cite{song2019generative} to minimize the loss function
        \begin{equation}
            \mathop{\mathbb{E}_{\tau, t, \eta}}\left[\left\Vert \frac{D_\theta(\tau + \eta, o, t)-\tilde{\tau}}{t^2} - \nabla_{\tau + \eta}\log q(\tau + \eta | \tau)\right\Vert_2^2\right]
            \label{eq:loss}
        \end{equation}
        with Gaussian noise $\eta \sim \mathcal{N}(0, t^2I)$ and a noise distribution $q$.
        The values of $t \sim v$ are sampled from a distribution $v$, commonly a logistic distribution~\cite{reuss2023multimodal}.
        Assuming a Gaussian distribution $q$ and noised sample $\tilde{\tau} = \tau + \eta$, substituting \autoref{eq:denoiser} in \autoref{eq:loss} results in the loss
        \begin{equation}
            \mathop{\mathbb{E}_{\tau, t, \eta}}\left[\left\Vert\frac{D_\theta(\tilde{\tau}, o, t)-\tau}{t^2}\right\Vert_2^2\right].
            \label{eq:loss_simplified}
        \end{equation}
    
        During inference, new samples of distribution $p$ are generated by gradually denoising samples of a unit Gaussian by following the approximated score function.
        This is done by solving the Probability Flow \gls{ode} with common \gls{ode} solvers such as Euler's method or solvers that are specifically designed for fast inference in diffusion~\cite{lu2022dpm}.
        
        Training for \gls{il} in robotics involves using observations $o$ and corresponding action sequences $\tau$ to learn an approximation of the data distribution's score function.
        During inference, new action sequences are generated by initially taking a random sample from a standard Gaussian distribution.
        This sample is refined through the Probability Flow \gls{ode} using an \gls{ode} solver.
        This process involves iteratively updating the sample by passing it to the model conditioned on the observation.

    \subsection{Movement Primitive Diffusion}
    \label{sec:methods:mpd}
         We propose \gls{mpd}, to combine the advantages of \glspl{sgm} and \glspl{prodmp}.
         In \gls{mpd}, the inner model $F_\Theta$ of \autoref{eq:scaler_main} consists of a trainable model $E_\Theta$ that outputs a weight vector $\bm w$.
         Combined with initial values $s_0$ for position and velocity, $\bm w$ is decoded into an action sequence $\tau$ using a \gls{prodmp}.
         Conceptually, model $E_\Theta$ denoises an action sequence conditioned on observations and maps it into the \gls{prodmp} weight space.
         The \gls{prodmp} acts as a decoder model to map the denoised weights back into action sequence space.
         The architecture of $F_\Theta$ is illustrated in \autoref{fig:overview}.
         The preconditioning functions of \autoref{eq:scaler_main} match the context of action generation with \glspl{mp}. 
         In Diffusion Policy, the network output is the noise that is removed from the sample in one diffusion step.
         For BESO, the network output is a fusion of absolute sample values and noise.
         However, for \gls{mpd}, the inner model output is generated by a \gls{prodmp} that directly outputs trajectory values, so $c_{\text{skip}}(t) = 0$ and $c_{\text{out}}(t) = 1$ are adapted accordingly.
         The parameters 
        \begin{equation}
            \begin{aligned}
                c_{\text{noise}}(t) &=\log(t)/4, &&
                c_{\text{in}}(t) &= 1/\sqrt{t^2+\sigma_{d}^2}
            \end{aligned}
            \label{eq:mpd_preconditioning}
        \end{equation}
        with $\sigma_d=0.5$ are identical to the values in~\cite{karras2022elucidating} and~\cite{reuss2023multimodal}.
        
        Robot control benefits from utilizing fine-grained high-frequency action sequences.
        In \gls{dil}, the input and output sizes of the learned model are determined by the number of elements in the denoised action sequence.
        Using high-frequency action sequences for a fixed time window has a substantial impact on model size.
        To manage this complexity, action sequences are commonly predicted at a lower frequency, followed by upsampling through interpolation techniques or \gls{mpc}~\cite{chi2023diffusionpolicy}.
        However, \gls{mpd} uses \glspl{prodmp} that can predict action sequences at arbitrary frequencies.
        \Gls{mpd} diffuses a low-frequency action sequence to shape the encoded \gls{prodmp} weights.
        At the final diffusion step, the encoded weight vector is decoded into a high-frequency action sequence for fine-grained robot control.
        This allows for fewer values in the action sequences during diffusion and further decouples the frequency of demonstration data from predicted action sequences.
        
        A naive alternative to \gls{mpd} is to diffuse action sequences in the \gls{prodmp}'s weight space to approximate the score function of weight vector distribution $v(w | o, \sigma(t))$ instead of action sequence distribution $p(\tau | o, \sigma(t))$.
        In preliminary experiments, we found that diffusing in weight space does not reach the high success rates of diffusing action sequences directly.
        Diffusing action sequences can leverage sophisticated network architectures such as transformers that benefit from the sequential nature of action sequences, which are not explicit in weight space.
        \Gls{mpd} thus focuses on diffusing action sequences directly.

        \gls{mpd} generates smooth, multimodal, high-frequency action sequences with guaranteed initial conditions for position and velocity.
        Utilizing \gls{prodmp} helps modeling temporal correlations between actions, which increases data efficiency and generates motions that are suitable for gentle manipulation of deformable objects.
        \gls{mpd} builds on the \gls{sgm} framework and learns to estimate the score function of the data distribution that governs the demonstration data.
    
\section{Experiments}
\label{sec:experiments}

    \subsection{Tasks}
    \label{sec:experiments:tasks}
    \def\mypicturesize{2.0cm}
\begin{figure}
    \centering
    \vspace{1.5mm}
    \begin{tikzpicture}[align=center]
        \node[inner sep=1pt, anchor=south west] (GLT start) {\includegraphics[height=\mypicturesize]{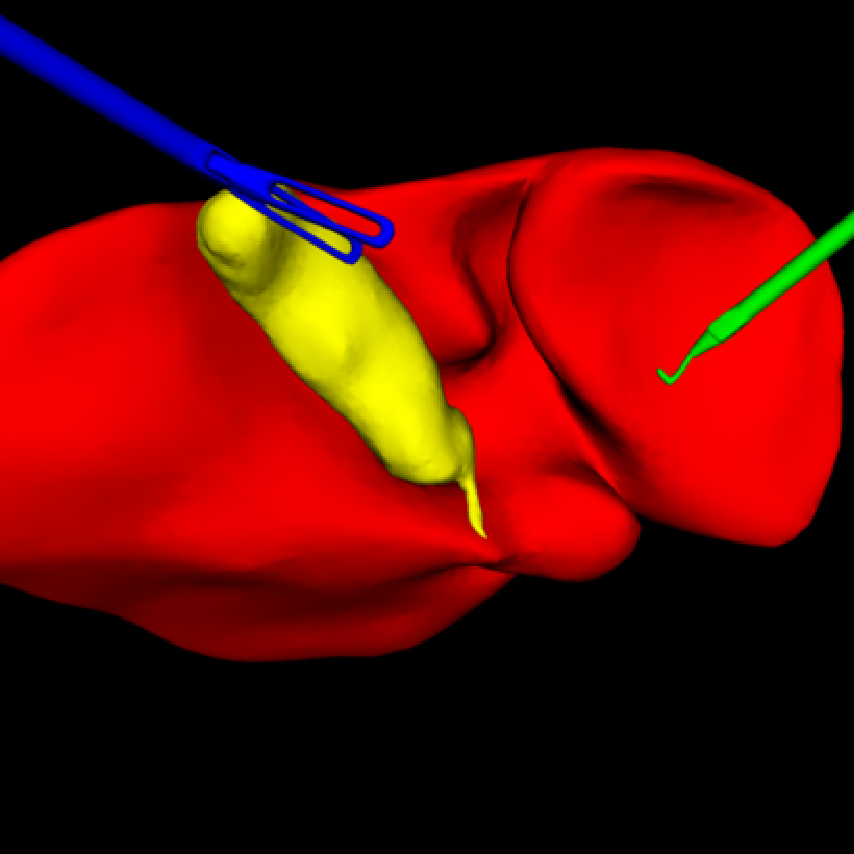}};
        \node[inner sep=1pt, anchor=south west, right = 0 of GLT start] (GLT mid) {\includegraphics[height=\mypicturesize]{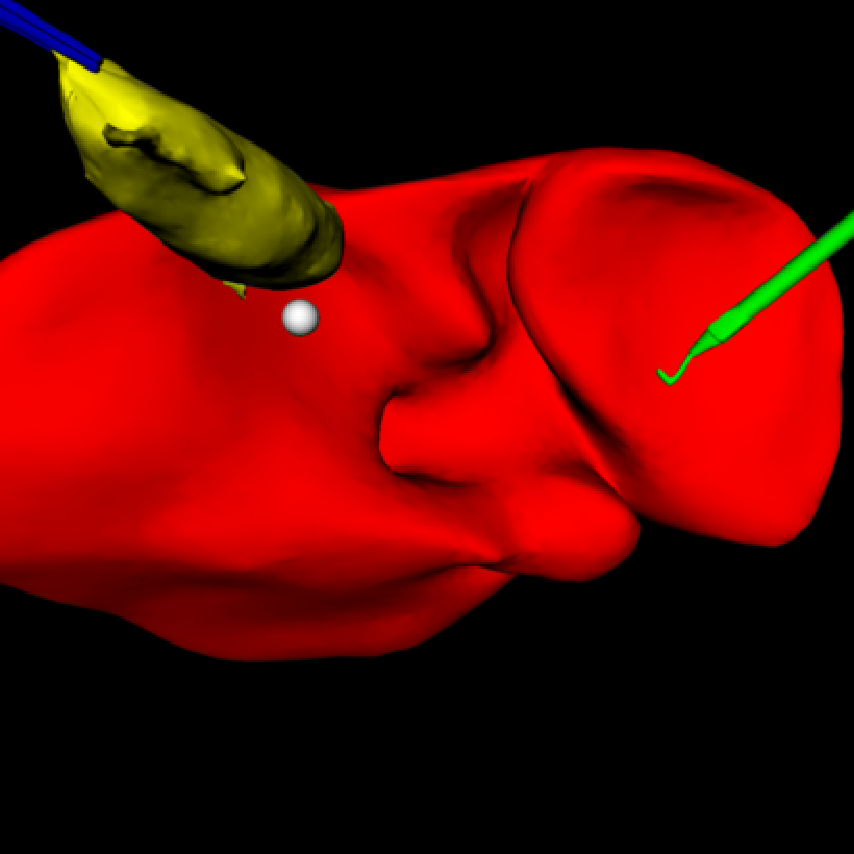}};
        \node[inner sep=1pt, anchor=south west, right = 0 of GLT mid] (GLT end) {\includegraphics[height=\mypicturesize]{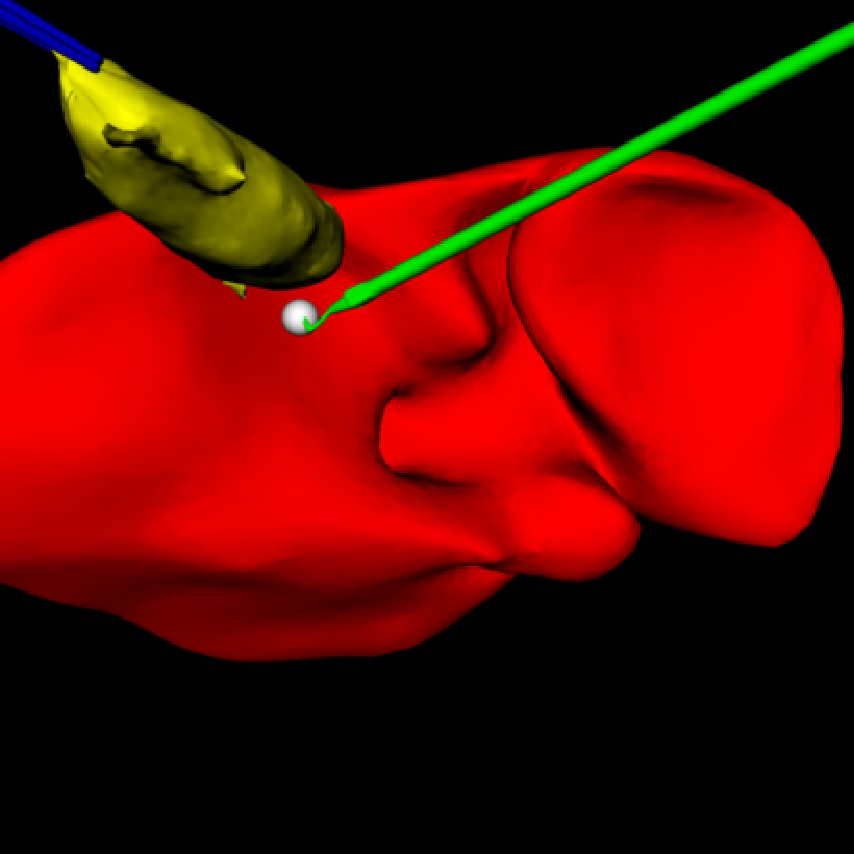}};
        \node[inner sep=1pt, anchor=south west, right = 0 of GLT end] (GLT RW) {\includegraphics[height=\mypicturesize]{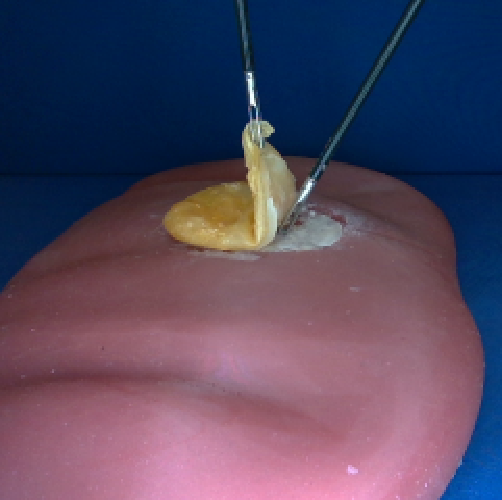}};
        \node[rotate=90, anchor=south, outer sep=0pt, inner sep=0pt] at ($(GLT start.west) + (-0.7mm, 0)$) {\small GLT};
        
        \node[inner sep=1pt, anchor=south west, below = 0 of GLT start] (RT start) {\includegraphics[height=\mypicturesize]{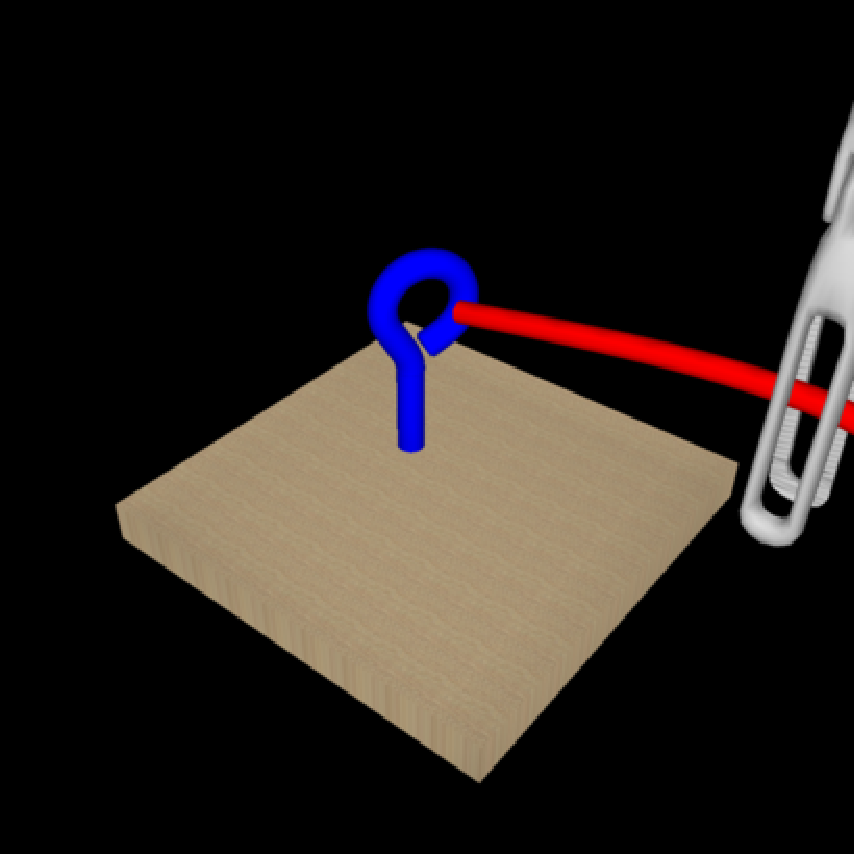}};
        \node[inner sep=1pt, anchor=south west, right = 0 of RT start] (RT mid) {\includegraphics[height=\mypicturesize]{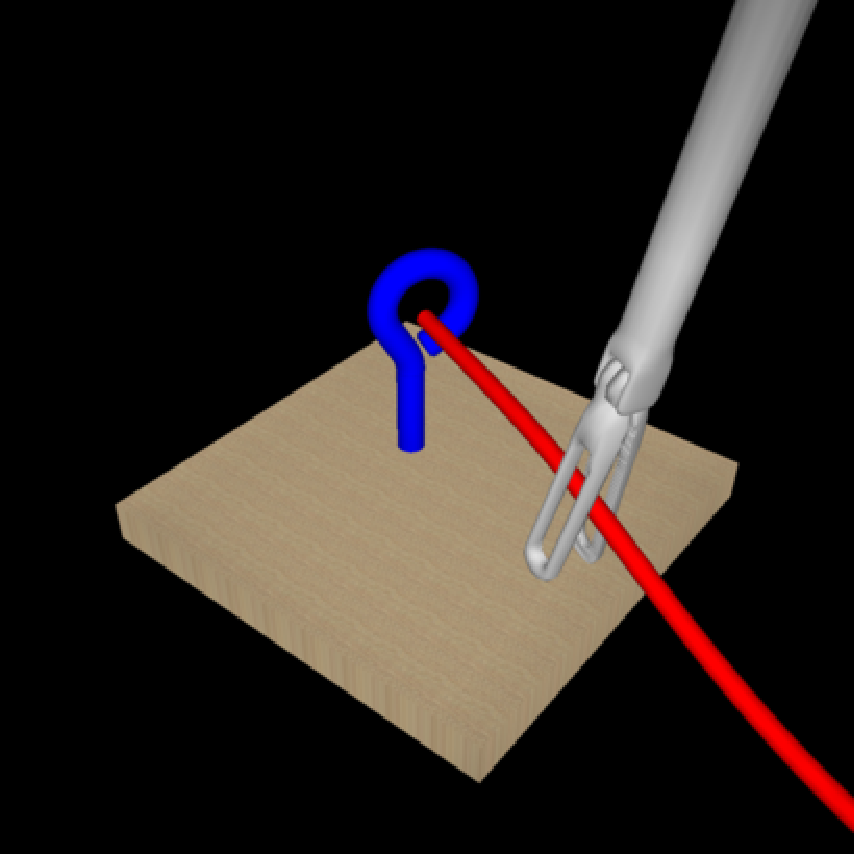}};
        \node[inner sep=1pt, anchor=south west, right = 0 of RT mid] (RT end) {\includegraphics[height=\mypicturesize]{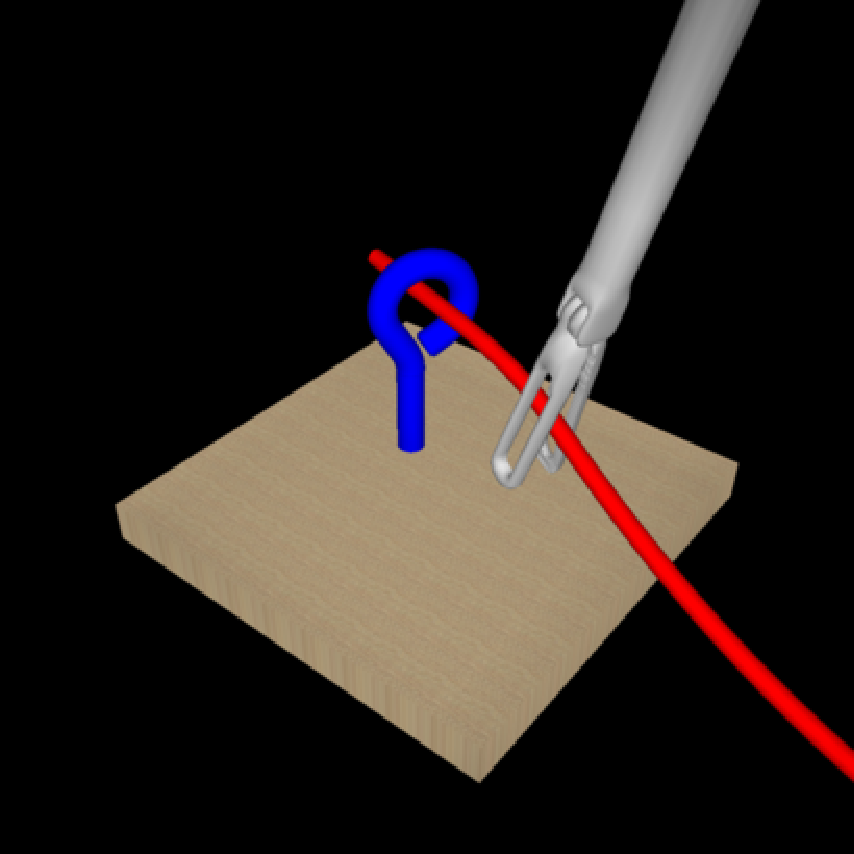}};
        \node[inner sep=1pt, anchor=south west, right = 0 of RT end] (RT RW) {\includegraphics[height=\mypicturesize]{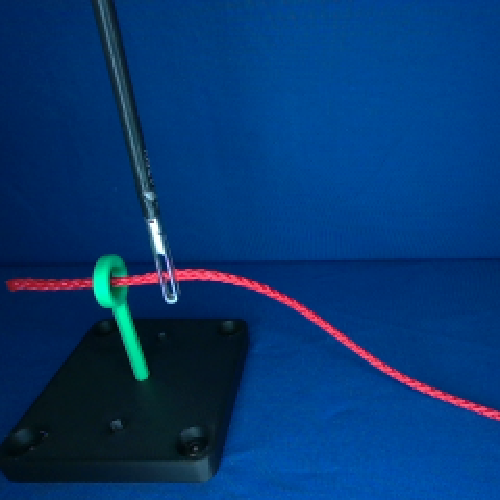}};
        \node[rotate=90, anchor=south, outer sep=0pt, inner sep=0pt] at ($(RT start.west) + (-0.7mm, 0)$) {\small RT};

        \node[inner sep=1pt, anchor=south west, below = 0 of RT start] (BTM start) {\includegraphics[height=\mypicturesize]{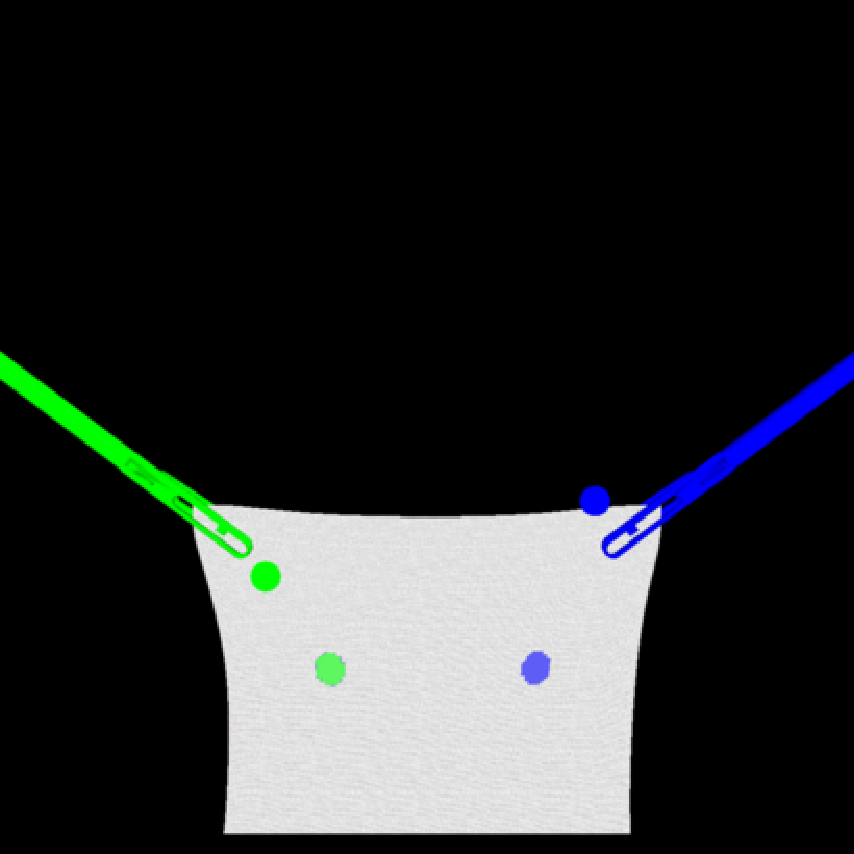}};
        \node[inner sep=1pt, anchor=south west, right = 0 of BTM start] (BTM mid) {\includegraphics[height=\mypicturesize]{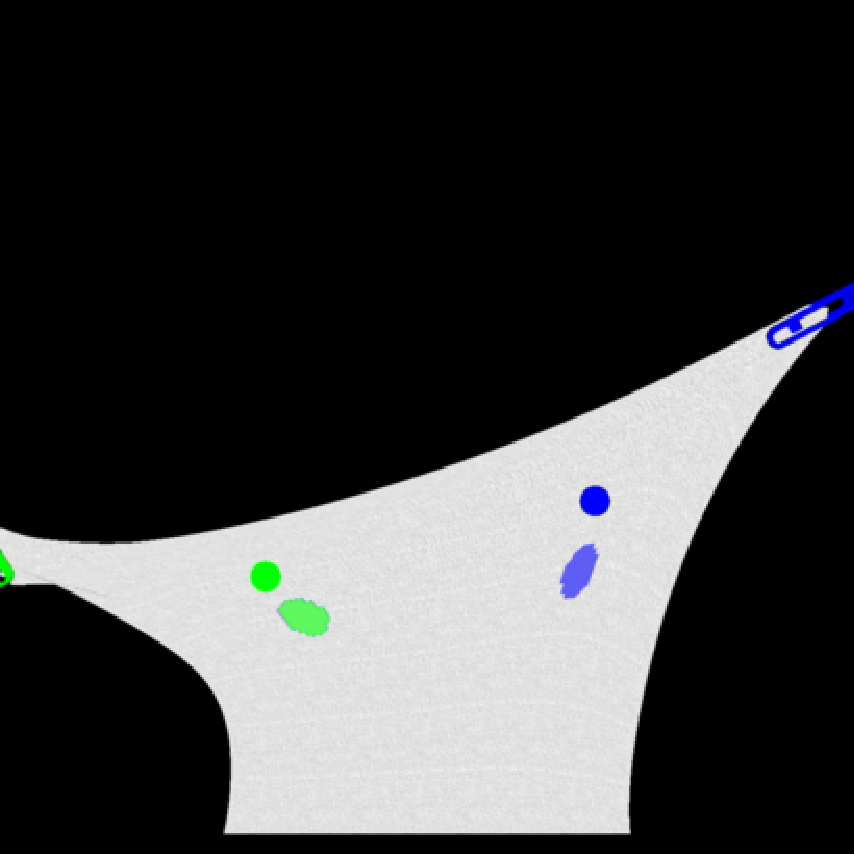}};
        \node[inner sep=1pt, anchor=south west, right = 0 of BTM mid] (BTM end) {\includegraphics[height=\mypicturesize]{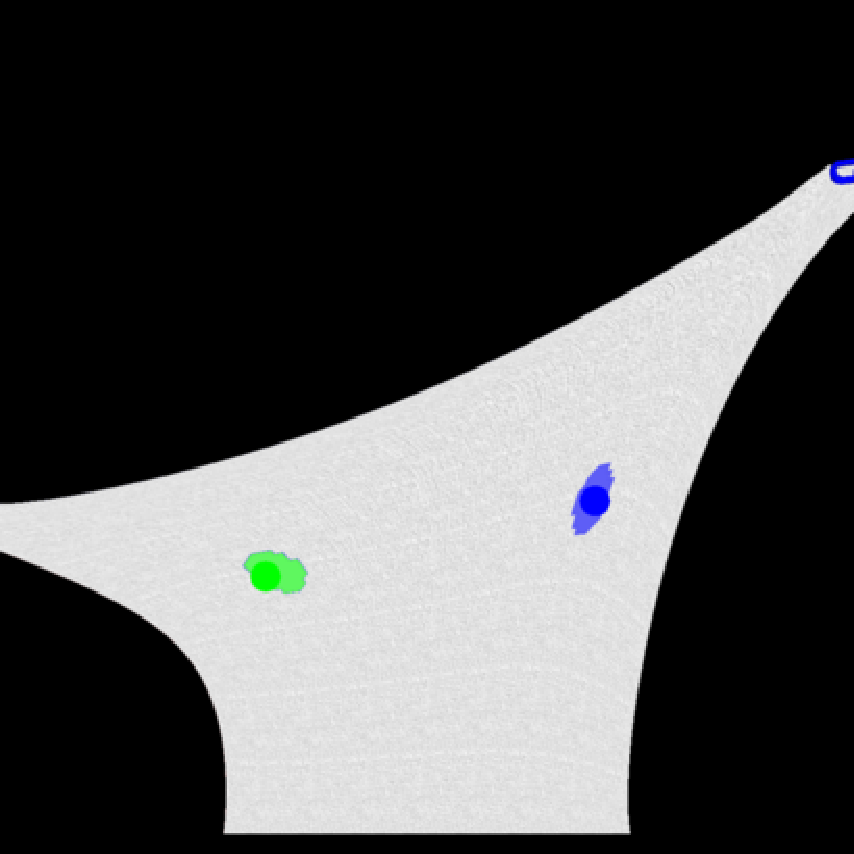}};
        \node[inner sep=1pt, anchor=south west, right = 0 of BTM end] (BTM RW) {\includegraphics[height=\mypicturesize]{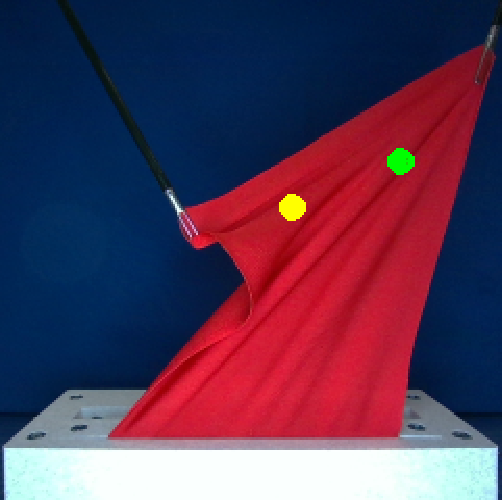}};
        \node[rotate=90, anchor=south, outer sep=0pt, inner sep=0pt] at ($(BTM start.west) + (-0.7mm, 0)$) {\small BTM};

        \node[inner sep=1pt, anchor=south west, below = 0 of BTM start] (LL start) {\includegraphics[height=\mypicturesize]{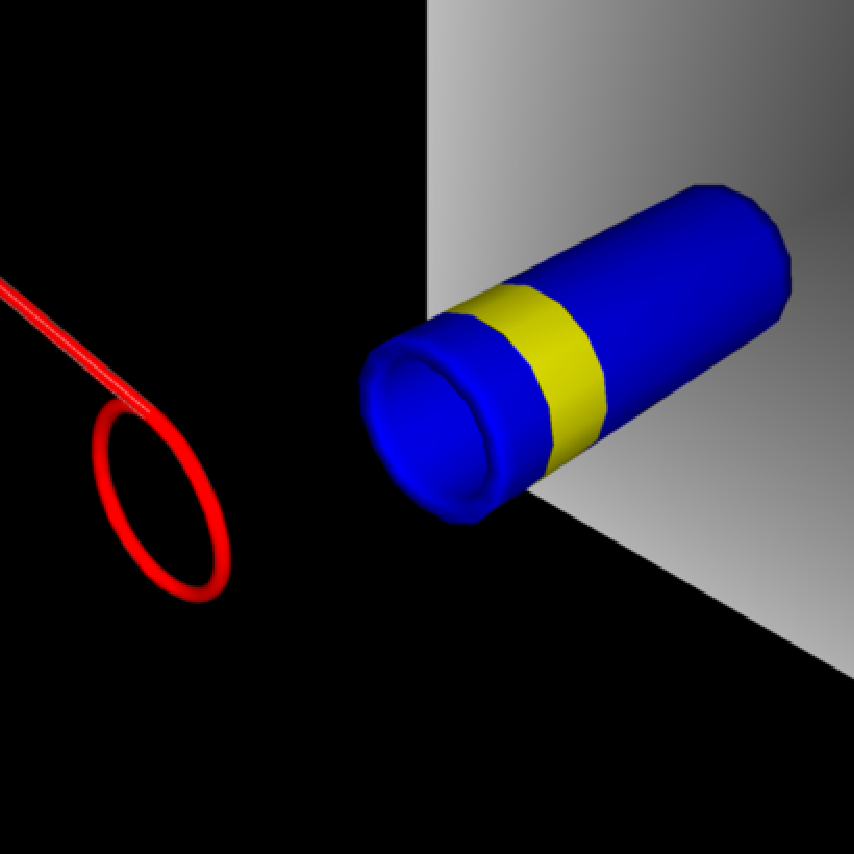}};
        \node[inner sep=1pt, anchor=south west, right = 0 of LL start] (LL mid) {\includegraphics[height=\mypicturesize]{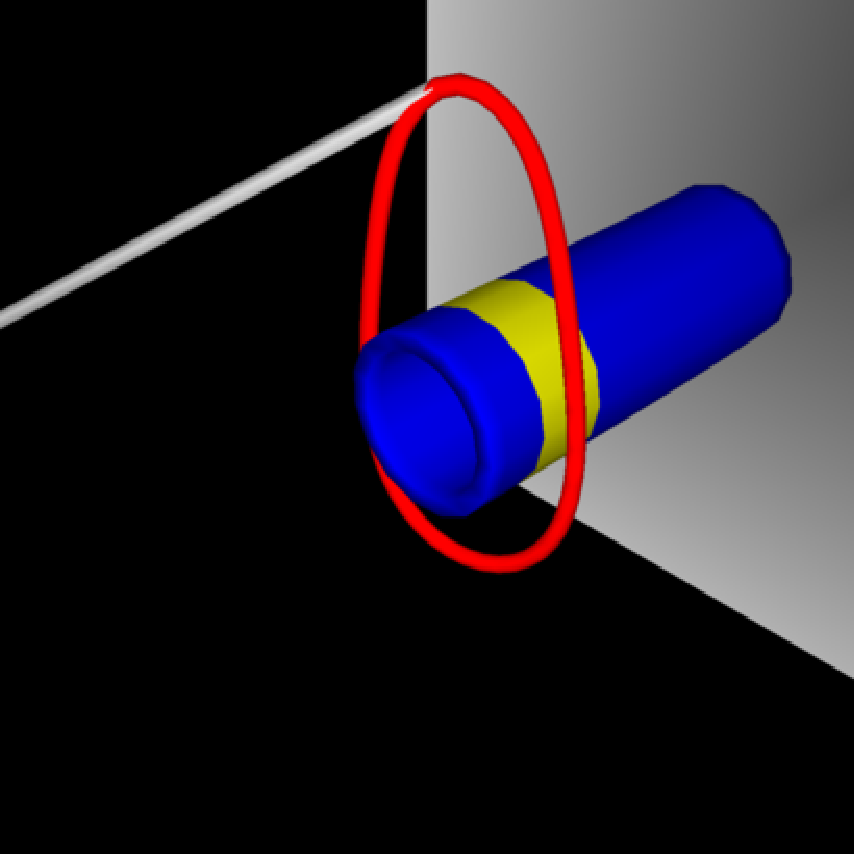}};
        \node[inner sep=1pt, anchor=south west, right = 0 of LL mid] (LL end) {\includegraphics[height=\mypicturesize]{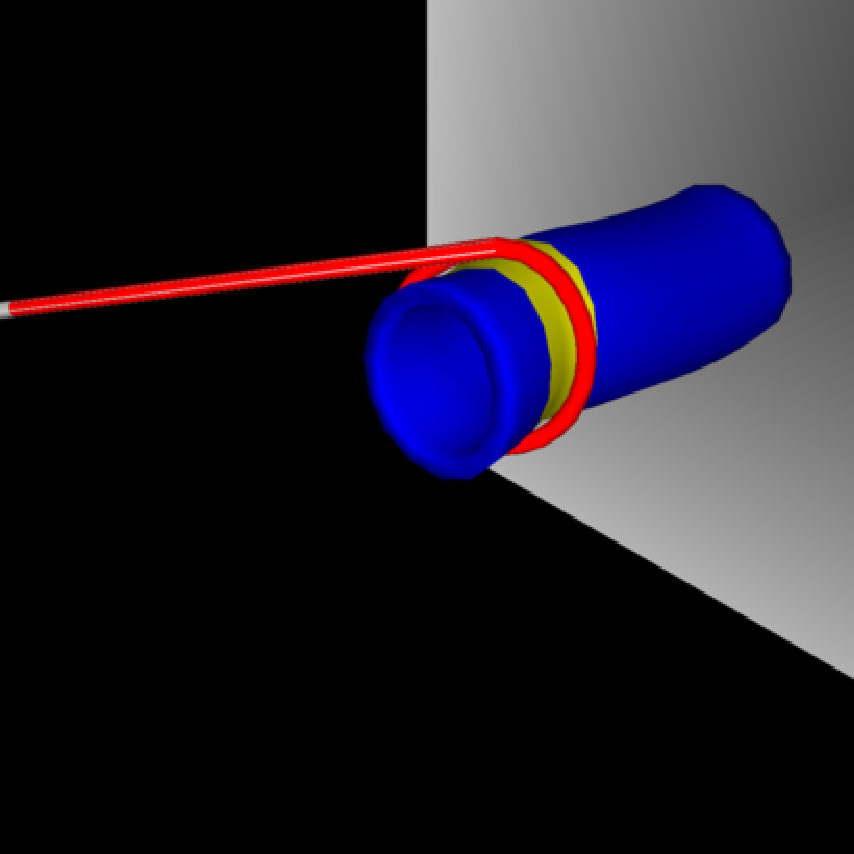}};
        \node[inner sep=1pt, anchor=south west, right = 0 of LL end] (LL RW) {\includegraphics[height=\mypicturesize]{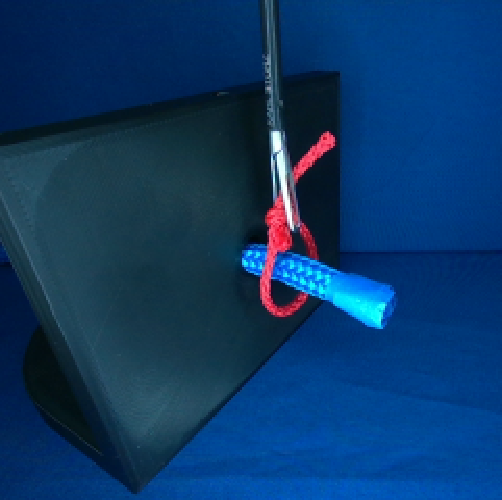}};
        \node[rotate=90, anchor=south, outer sep=0pt, inner sep=0pt] at ($(LL start.west) + (-0.7mm, 0)$) {\small LL};
        
    \end{tikzpicture}
    \caption{Start, intermediate, and end state of the tasks in simulation. The final column shows the respective real world experiment. Grasp Lift Touch (GLT) requires sequential collaboration between instruments, Rope Threading (RT) and Ligating Loop (LL) depend on accurate alignment deformable ropes, and Bimanual Tissue Manipulation (BTM) requires concurrent collaboration between instruments to control the shape of a deformable tissue.}
    \label{fig:tasks}
\end{figure}
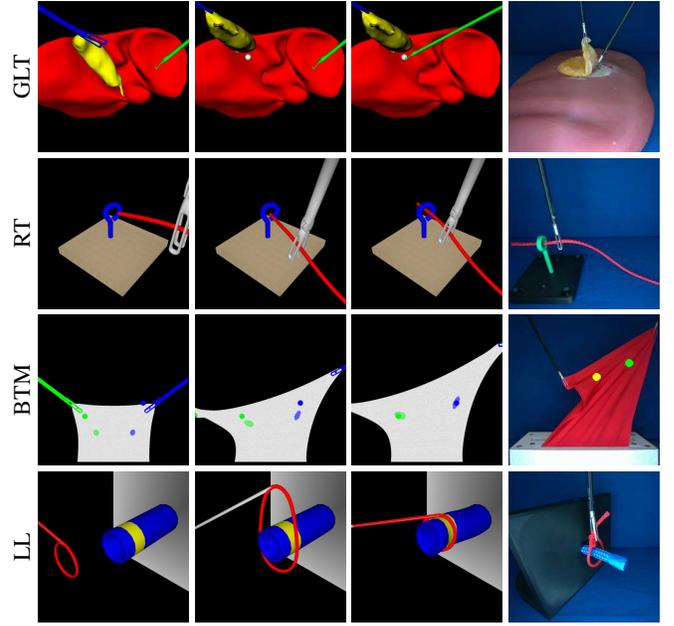

    \begin{figure}
    \centering
    \vspace{1.5mm}
    \begin{tikzpicture}
        \node[inner sep=1pt] (rw1) {\includegraphics[width=2.7cm]{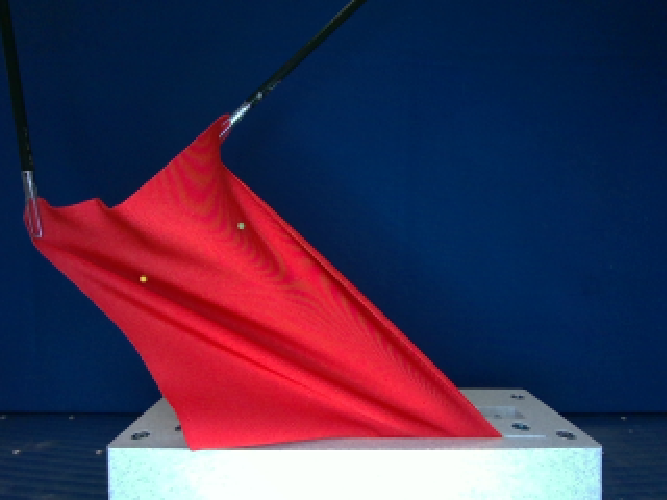}};
        \node[inner sep=1pt, anchor=west] (rw2) at (rw1.east) {\includegraphics[width=2.7cm]{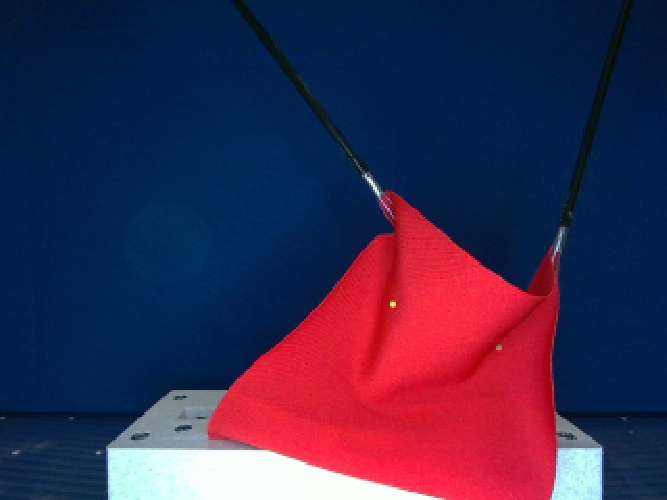}};
        \node[inner sep=1pt, anchor=west] (rw3) at (rw2.east) {\includegraphics[width=2.7cm]{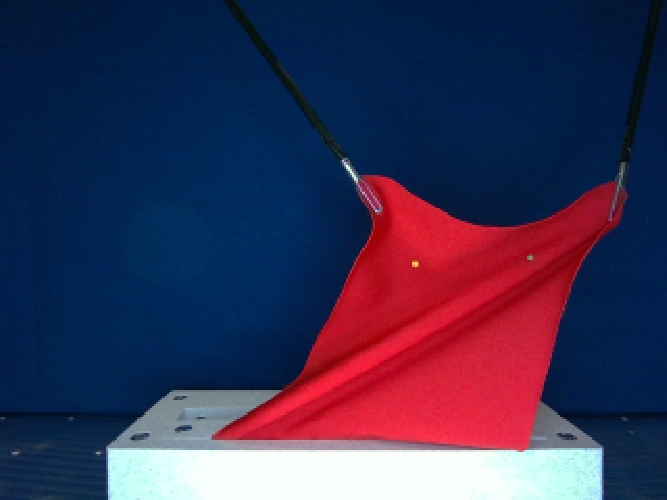}};
    \end{tikzpicture}
    \caption{
        Example images from the real world task.
        The deformation behavior of the tissue differs significantly from the simulated variant.
        When stretched, it throws folds and has large variability when slacking without tension, \eg, bulging forward or folding in. 
        }
    \label{fig:rw_btm}
\end{figure}

    In our experiments, we evaluate how \gls{mpd} performs to align with requirements for application in \gls{ras}, based on success rate, motion quality, and data efficiency.
    \gls{mpd} is evaluated on four different simulated LapGym~\cite{scheikl2023lapgym} tasks and their respective real world robotic setups, illustrated in \autoref{fig:tasks}.
    The tasks represent different types of motion such as cooperation of instruments, grasping, and deformable object manipulation, all of which are crucial for successful application in \gls{ras}.
    
    In \gls{glt}, two laparoscopic instruments are controlled to first grasp a gallbladder, lift the gallbladder to expose a target point in the liver's gallbladder bed, and finally touch the target point with an electrocautery hook.
    The task requires sequential coordination of the instruments as well as grasping and manipulating deformable objects.
    The manipulation of the gallbladder is a straightforward retraction motion.
    The exact deformed shape of the gallbladder plays a subordinate role in exposing the target point as long as the grasped point is sufficiently pulled back.
    
    In \gls{btm}, two laparoscopic graspers are attached to the corners of a piece of tissue that is marked with two points.
    The graspers are controlled to deform the tissue in a way that the two marked points overlap with target points in Cartesian space.
    In contrast to retraction tasks, this indirect tissue manipulation requires control of the complete shape of the deformable object, and coordination between the instruments is continuous, as motion of any instrument influences the shape of the whole tissue.
    Both \gls{glt} and \gls{btm} are characterized by motions that are determined by the instrument's current and target positions.
    
    In \gls{rt}, a rope is held by a laparoscopic grasper that is controlled to thread the rope through an eyelet.
    In contrast to the previous tasks, \gls{rt} cannot be solved by directly moving the instrument to the target position, but instead has to match waypoints that ensure that the tip of the rope is correctly threaded through the eyelet.
    
    Finally, \gls{ll} presents a similar challenge but adds further complexity by requiring additional indirect manipulation of a deformable object.
    In this task, the instrument consists of a rigid shaft and a deformable loop that can be closed to constrict tubular objects, such as blood vessels.
    The instrument shaft is controlled to maneuver the loop over a tube and constrict the loop around a colored marking.
    Compared to \gls{rt}, the loop is much more flexible, deforms more under instrument motion, and requires careful control to navigate a bottleneck of viable waypoints.

    The real world tasks replicate the simulated tasks with a setup of xArm7 manipulators (UFactory, China) and laparoscopic graspers (KARL STORZ SE \& Co. KG, Germany).
    RGB images are captured using a D405 camera (Intel RealSense, USA).
    The spatial resolution from image pixels to Cartesian positions is approximately \SI{1}{\mm}.
    The real world setup of \gls{ll} is simplified by omitting the constriction step of the loop.
    A silicone phantom of a porcine liver with an attached latex gallbladder is used for the \gls{glt} setup.
    The real world setup of \gls{btm} uses a piece of $84\,\%$ polyamid $16\,\%$ elasthan cloth as tissue and exhibits more complex deformations compared to the simulated task (see \autoref{fig:rw_btm}).

    \gls{mpd} is evaluated against three baselines, namely BESO~\cite{reuss2023multimodal}, and two variants of Diffusion Policy~\cite{chi2023diffusionpolicy}, DP-C and DP-T.
    BESO and DP-T are based on transformers, while DP-C employs a 1D temporal CNN model architecture.
    
    \subsection{Evaluation Metrics}
    \label{sec:experiments:metrics}
        \textbf{Motion Metrics:} 
        Applications in \gls{ras} have specific requirements for motion behavior in addition to raw success rate of task completion.
        Tissue acceleration is a metric to quantify surgical performance for gentle manipulation of delicate tissue~\cite{sugiyama2022tissue} and should be minimized to reduce risk of tissue damage.
        We further characterize gentle motions by smooth instrument trajectories.
        We quantify smoothness based on minimizing instrument jerk to increase safety during motion execution and reduce wear of mechanical components.
        The motion of skilled surgeons is furthermore characterized by efficient movements~\cite{hove2010objective}.
        In our work, efficiency is measured by three features, all of which should be minimized: instrument energy as the sum of accelerations over task execution, path length as the travelled distance of the instrument, and episode length as the time to task completion.
        The five metrics tissue acceleration, instrument jerk, instrument energy, path length, and episode length are evaluated for all tasks, with the following task specific adaptations to represent the metrics.
        For \gls{rt}, tissue acceleration is measured as the acceleration of points on the rope.
        For \gls{ll}, the instrument consists of a rigid shaft and a deformable loop, so instrument jerk is examined for both parts individually.
        On the real world tasks, the full state of the tissue is not accessible, so tissue acceleration cannot be measured directly.
        However, for \gls{btm}-RW, marker acceleration is tracked as a surrogate.
        
        \textbf{Data Efficiency:}
        Data scarcity in the medical domain, particularly in specialized fields such as surgery, requires the development of machine learning methods that efficiently learn from limited data, ensuring reliable applications in clinical settings.
        All methods are trained multiple times using varying numbers of demonstrations to evaluate their data efficiency with respect to achieved success rate.
        
    \subsection{Hyper Parameters}
    \label{sec:methods:hyper}
        \textbf{Architecture:}
        For all methods, including the baselines, the model architecture parameters are individually optimized during preliminary experiments on \gls{btm} as the task is evaluated in simulation and the real world.
        We found that DP-C performs best with the original values reported in~\cite{chi2023diffusionpolicy} with CNN layer sizes of $(256, 512, 1024)$.
        BESO, DP-T, and \gls{mpd} share an optimal transformer architecture, proposed in~\cite{chi2023diffusionpolicy}, with $6$ layers, $4$ heads, attention dropout probability of $0.3$ and embedding size of $256$.
        
        \textbf{Observations:}
        All methods are trained for both state and raw RGB image observations.
        State observations include task-specific state vectors defined in LapGym~\cite{scheikl2023lapgym}.
        Image observations are $256 \times 256$ RGB images that are randomly cropped to $224 \times 224$ during training and center-cropped during inference.
        Images are encoded by an adapted ResNet-18 architecture, as described in~\cite{chi2023diffusionpolicy}, without pretraining and an output size of $128$.
        For the transformer-based models, the $m$ image observations are encoded independently.
        For DP-C, the $m$ RGB images are stacked along the color channel before encoding.

        \textbf{Action Sequences:}
        We predict action sequences of length $n=12$, conditioned on the previous $m=3$ observations.
        The time $\Delta T$ between both successive observations $o_j$ and successive actions $\tau_i$ is \SI{0.1}{\s}.
        For execution on the real world task, we further generate high-frequency action sequences with $\Delta T = \SI{0.005}{\s}$.
        The baseline methods require upsampling to calculate the high-frequency action sequences.
        Linear interpolation is selected for simplicity and computational effectiveness.
        \gls{mpd} does not require upsampling, as generating high-frequency action sequences is directly supported by \glspl{prodmp}.
        
        \textbf{Diffusion:}
        Diffusion hyper parameters for the baselines, such as beta schedule and \gls{ode} solver, are set according to the respective works.
        Through initial experiments, we found that Diffusion Policy requires significantly fewer diffusion steps during inference than used during training.
        We train Diffusion Policy with $100$ diffusion steps, and use $5$ and $10$ diffusion steps during inference for DP-T and DP-C, respectively, without loss of performance.
        This greatly improves execution time, as only a fraction of model forward passes are required during inference.
        For \gls{mpd} we use Euler's method as the \gls{ode} solver, as more sophisticated samplers did not show noticeable improvements.
        For the \gls{prodmp} we use $3$ basis functions.

        \textbf{Training:}
        The number of training demonstrations in the datasets are $|\mathcal{D}_{\text{GLT}}| = 90$, $|\mathcal{D}_{\text{RT}}| = 200$, $|\mathcal{D}_{\text{BTM}}| = 150$, $|\mathcal{D}_{\text{LL}}| = 135$, $|\mathcal{D}_{\text{BTM-RW}}| = 200$, and $|\mathcal{D}_{\text{GLT-RW}}| = |\mathcal{D}_{\text{RT-RW}}| = |\mathcal{D}_{\text{LL-RW}}| = 45$.
        All demonstrations are captured at \SI{10}{\Hz} by a single proficient human demonstrator with an Xbox One controller (Microsoft Corp., USA).
        \gls{mpd} and the baseline methods are evaluated every $100$ training epochs for $100$ task executions.
        Each method is trained on $r=5$ random seeds for $3\,000$ epochs.
        The motion metrics defined in \autoref{sec:experiments:metrics} are aggregated on the individual training run's epoch that achieved the highest success rate to allow for an accurate comparison of each method's capability by considering its best-case performance scenario.
        
        We use AdamW~\cite{loshchilov2018adamw} as optimizer with learning rate, betas, and weight decay as described in~\cite{chi2023diffusionpolicy}.
        The real world tasks are evaluated using a fully trained model on $50$ task executions.

\section{RESULTS}
\label{sec:results}
    \textbf{Success Rate:}
    \autoref{tab:successrate} reports the success rates for all methods evaluated on the simulation tasks.
    \begin{table}
    \centering
    \vspace{1.5mm}
    \caption{
        Success rate mean and standard deviation across all simulation tasks based on $5$ trained models and $100$ rollouts.
        The best method is \textbf{bold}, the second best \underline{underlined}.
    }
    \label{tab:successrate}
    \begin{tabular}{l c c c c c}
        \toprule
         & \textbf{GLT} & \textbf{RT} & \textbf{BTM} & \textbf{LL} & \textbf{Average} \\
        \midrule
        \multicolumn{6}{c}{\textbf{State Observations}} \\
        BESO & 68.4 (6.8) & \underline{90.4} (1.2) & 88.2 (3.9) & 83.2 (4.8) & 82.55 \\
        DP-T & \textbf{100} (0.0) & 89.6 (1.9) & \underline{98.4} (0.5) & \textbf{100} (0.0) & \underline{97.00} \\
        DP-C & \textbf{100} (0.0) & 82.0 (2.3) & 93.2 (1.9) & \textbf{100} (0.0) & 93.80 \\
        MPD & \underline{99.2} (0.7) & \textbf{93.8} (1.2) & \textbf{99.0} (0.6) & \underline{99.6} (0.5) & \textbf{97.90} \\
        \midrule
        \multicolumn{6}{c}{\textbf{Image Observations}} \\
        BESO & \underline{99.8} (0.4) & 66.8 (2.3) & 91.4 (2.4) & \underline{99.8} (0.4) & 89.45 \\
        DP-T & \textbf{100} (0.0) & 76.6 (4.5) & \underline{95.8} (0.7) & \underline{99.8} (0.4) & \underline{93.05} \\
        DP-C & \textbf{100} (0.0) & \textbf{83.2} (1.2) & 85.2 (1.3) & \textbf{100} (0.0) & 92.15 \\
        MPD & \textbf{100} (0.0) & \underline{78.6} (3.4) & \textbf{99.0} (1.1) & \textbf{100} (0.0) & \textbf{94.40} \\
        \bottomrule
    \end{tabular}
\end{table}

    \gls{mpd} outperforms the baseline methods with average success rates of $97.9\,\%$ and $94.4\,\%$ for state and image observations.
    DP-T is the second best, reaching $97.0\,\%$ and $93.05\,\%$, respectively.
    \gls{mpd} and DP-T further show more consistent performance over different runs compared to the other methods.
    BESO consistently reaches higher success rates on image observations compared to state observations, while the other methods perform better on state observations.
    Yet, BESO is still outperformed by the other methods on image observations.
     
    On the real world task, only \gls{mpd} and DP-C are evaluated.
    Even though DP-T and BESO outperform DP-C on the simulated \gls{btm} task, we do not evaluate them on the real world tasks because the generated action sequences were not gentle enough for execution on the real robot without significant post-processing.
    \autoref{fig:trajectories} illustrates generated trajectories of all methods on the simulated \gls{btm} task.
    \def\plotfraction{0.45}
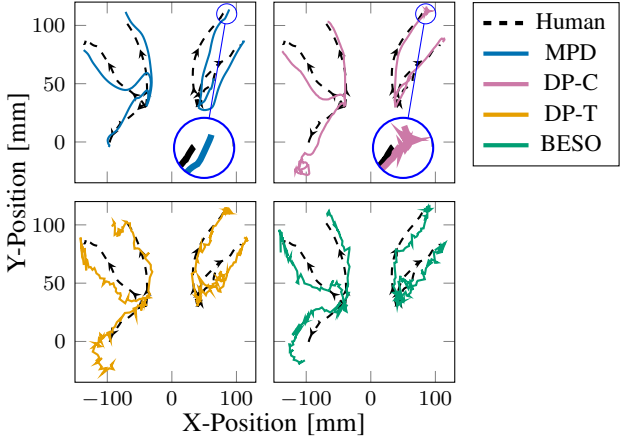
\begin{figure}
\centering
\vspace{1.5mm}

\begin{tikzpicture}[spy using outlines={circle, magnification=3, connect spies}]
\begin{axis}[
    name=MPDtraj,
    ymin=-35,
    ymax=120,
    xmin=-150,
    xmax=130,
    width=\plotfraction\columnwidth,
    height=\plotfraction\columnwidth,
    ylabel={Y-Position [mm]},
    ylabel style={at={(-0.2, 0.0)},},
    x tick label style={font=\footnotesize},
    y tick label style={font=\footnotesize},
    legend style={
        at={(2.2,1.0)},
        anchor=north west,
        legend columns=1,
        font=\small,
    },
    legend entries={Human, MPD, DP-C, DP-T, BESO},
    xticklabels={,,},
]
\addlegendimage{ultra thick, dashed}
\addlegendimage{ultra thick, color=MPD}
\addlegendimage{ultra thick, color=DPC}
\addlegendimage{ultra thick, color=DPT}
\addlegendimage{ultra thick, color=BESO}

\foreach \n in {1,2,3}{
    \addplot[dashed, thick, mark=none, color=black, decoration={markings, mark=between positions 0.2 and 0.95 step 1.5em with {\arrow [scale=0.8]{stealth}}}, postaction=decorate] table [x=LX, y=LY, col sep=comma] {data/trajectories/HUMAN/tool_trajectory_\n.csv};
    \addplot[dashed, thick, mark=none, color=black, decoration={markings, mark=between positions 0.2 and 0.95 step 1.5em with {\arrow [scale=0.8]{stealth}}}, postaction=decorate] table [x=RX, y=RY, col sep=comma] {data/trajectories/HUMAN/tool_trajectory_\n.csv};
}
\foreach \n in {1,2,3}{
    \addplot[thick, mark=none, color=MPD] table [x=LX, y=LY, col sep=comma] {data/trajectories/MPD/tool_trajectory_\n.csv};
    \addplot[thick, mark=none, color=MPD] table [x=RX, y=RY, col sep=comma] {data/trajectories/MPD/tool_trajectory_\n.csv};
}

  \coordinate (spypoint0) at (axis cs:85,110);
  \coordinate (magnifyglass0) at (axis cs:50,-5);

\end{axis}
\spy [blue, size=0.8cm] on (spypoint0) in node[fill=white] at (magnifyglass0);

\begin{axis}[
    name=DPCtraj,
    at=(MPDtraj.south east),
    ymin=-35,
    ymax=120,
    xmin=-150,
    xmax=130,
    anchor=left of south west,
    width=\plotfraction\columnwidth,
    height=\plotfraction\columnwidth,
    yticklabels={,,},
    xticklabels={,,},
]
\foreach \n in {1,2,3}{
    \addplot[dashed, thick, mark=none, color=black, decoration={markings, mark=between positions 0.2 and 0.95 step 1.5em with {\arrow [scale=0.8]{stealth}}}, postaction=decorate] table [x=LX, y=LY, col sep=comma] {data/trajectories/HUMAN/tool_trajectory_\n.csv};
    \addplot[dashed, thick, mark=none, color=black, decoration={markings, mark=between positions 0.2 and 0.95 step 1.5em with {\arrow [scale=0.8]{stealth}}}, postaction=decorate] table [x=RX, y=RY, col sep=comma] {data/trajectories/HUMAN/tool_trajectory_\n.csv};
}
\foreach \n in {1,2,3}{
    \addplot[thick, mark=none, color=DPC] table [x=LX, y=LY, col sep=comma] {data/trajectories/DPC/tool_trajectory_\n.csv};
    \addplot[thick, mark=none, color=DPC] table [x=RX, y=RY, col sep=comma] {data/trajectories/DPC/tool_trajectory_\n.csv};
}
  \coordinate (spypoint) at (axis cs:85,110);
  \coordinate (magnifyglass) at (axis cs:50,-5);
\end{axis}

\spy [blue, size=0.8cm] on (spypoint) in node[fill=white] at (magnifyglass);

\begin{axis}[
    name=DPTtraj,
    at=(MPDtraj.south west),
    yshift=-0.25cm,
    anchor=north west,
    ymin=-35,
    ymax=120,
    xmin=-150,
    xmax=130,
    width=\plotfraction\columnwidth,
    height=\plotfraction\columnwidth,
    xlabel={X-Position [mm]},
    xlabel style={at={(1.1, -0.12)},anchor=north},
    x tick label style={font=\footnotesize},
    y tick label style={font=\footnotesize},
]
\foreach \n in {1,2,3}{
    \addplot[dashed, thick, mark=none, color=black, decoration={markings, mark=between positions 0.2 and 0.95 step 1.5em with {\arrow [scale=0.8]{stealth}}}, postaction=decorate] table [x=LX, y=LY, col sep=comma] {data/trajectories/HUMAN/tool_trajectory_\n.csv};
    \addplot[dashed, thick, mark=none, color=black, decoration={markings, mark=between positions 0.2 and 0.95 step 1.5em with {\arrow [scale=0.8]{stealth}}}, postaction=decorate] table [x=RX, y=RY, col sep=comma] {data/trajectories/HUMAN/tool_trajectory_\n.csv};
}
\foreach \n in {1,2,3}{
    \addplot[thick, mark=none, color=DPT] table [x=LX, y=LY, col sep=comma] {data/trajectories/DPT/tool_trajectory_\n.csv};
    \addplot[thick, mark=none, color=DPT] table [x=RX, y=RY, col sep=comma] {data/trajectories/DPT/tool_trajectory_\n.csv};
}
\end{axis}
\begin{axis}[
    name=BESOtraj,
    at=(DPTtraj.south east),
    ymin=-35,
    ymax=120,
    xmin=-150,
    xmax=130,
    anchor=left of south west,
    width=\plotfraction\columnwidth,
    height=\plotfraction\columnwidth,
    yticklabels={,,},
    x tick label style={font=\footnotesize},
    y tick label style={font=\footnotesize},
]
\foreach \n in {1,2,3}{
    \addplot[dashed, thick, mark=none, color=black, decoration={markings, mark=between positions 0.2 and 0.95 step 1.5em with {\arrow [scale=0.8]{stealth}}}, postaction=decorate] table [x=LX, y=LY, col sep=comma] {data/trajectories/HUMAN/tool_trajectory_\n.csv};
    \addplot[dashed, thick, mark=none, color=black, decoration={markings, mark=between positions 0.2 and 0.95 step 1.5em with {\arrow [scale=0.8]{stealth}}}, postaction=decorate] table [x=RX, y=RY, col sep=comma] {data/trajectories/HUMAN/tool_trajectory_\n.csv};
}
\foreach \n in {1,2,3}{
    \addplot[thick, mark=none, color=BESO] table [x=LX, y=LY, col sep=comma] {data/trajectories/BESO/tool_trajectory_\n.csv};
    \addplot[thick, mark=none, color=BESO] table [x=RX, y=RY, col sep=comma] {data/trajectories/BESO/tool_trajectory_\n.csv};
}
\end{axis}

\end{tikzpicture}

\caption{
    Cartesian instrument positions of trajectories generated in reference to human demonstrations on the simulated Bimanual Tissue Manipulation task. In contrast to the baselines, MPD consistently generates smooth trajectories.
    }
\label{fig:trajectories}
\end{figure}

\begin{figure}
\centering
\begin{tikzpicture}
\begin{axis}[
    name=RWBTMSucc,
    width=1.0\columnwidth,
    height=0.5\columnwidth,
    xlabel={Distance Threshold [mm]},
    xlabel style={
        at={(0.5, -0.1)},
    },
    ylabel={Success Rate [\%]},
    ylabel style={
        at={(-0.08, 0.5)},
    },
    ytick={0,20,...,100},
    minor ytick={0,5,...,100},
    xtick={0,1,...,8},
    minor xtick={0,0.5,...,8},
    xmin=-0.3,
    xmax=8,
    legend style={
        at={(0.5,1.03)},
        anchor=south,
        legend columns=4,
        font=\small,
    },
    ymajorgrids=true,
    yminorgrids=true,
    major grid style={semithick, black!30!white, dotted},
    minor grid style={thin, black!20!white, dotted},
    minor y tick style = transparent,
    x tick label style={
        font=\footnotesize,
    },
    y tick label style={
        font=\footnotesize,
    },
    legend entries={MPD, MPD-Img, DP-C, DP-C-Img},
]
\addlegendimage{ultra thick, color=MPD}
\addlegendimage{ultra thick, color=MPD, dashed}
\addlegendimage{ultra thick, color=DPC}
\addlegendimage{ultra thick, color=DPC, dashed}

\addplot[color=MPD, very thick, solid] table[x=threshold,y=mpd_trajectories,col sep=comma] {data/success_rate_over_threshold.csv};
\addplot[color=MPD, very thick, dashed] table[x=threshold,y=mpd_img_trajectories,col sep=comma] {data/success_rate_over_threshold.csv};
\addplot[color=DPC, very thick, solid] table[x=threshold,y=dpu_trajectories,col sep=comma] {data/success_rate_over_threshold.csv};
\addplot[color=DPC, very thick, dashed] table[x=threshold,y=dpu_img_trajectories,col sep=comma] {data/success_rate_over_threshold.csv};

\coordinate (left_m) at (axis cs:4.3,28.0);
\coordinate (left_t) at (axis cs:4.1,40.0);
\coordinate (left_d) at (axis cs:3.8,30.0);

\coordinate (right_m) at (axis cs:5.6,48.0);
\coordinate (right_t) at (axis cs:5.4,58.0);
\coordinate (right_d) at (axis cs:5.9,57.0);

\coordinate (success) at (axis cs:4.0,5.0);

\end{axis}
\node[anchor=south east] (img) at ($(RWBTMSucc.south east) + (0, 0.1cm)$) {\includegraphics[height=1.8cm]{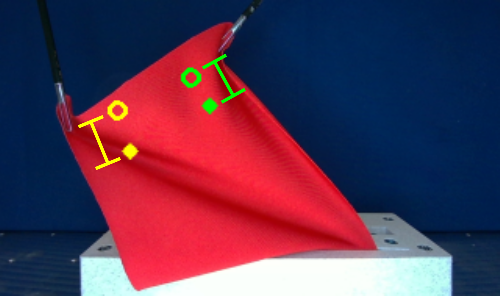}};
\node[green, anchor=north] at ($(img.north) + (0.15cm, -0.2cm)$) {$d_r$};
\node[yellow, anchor=west] at ($(img.west) + (0.15cm, 0)$) {$d_l$};

\end{tikzpicture}

\caption{
Success rate of state- and image-based policies on the real world \gls{btm} task in relation to the distance threshold $T$ between marker and target.
}
\label{fig:success_rate_over_threshold}
\end{figure}
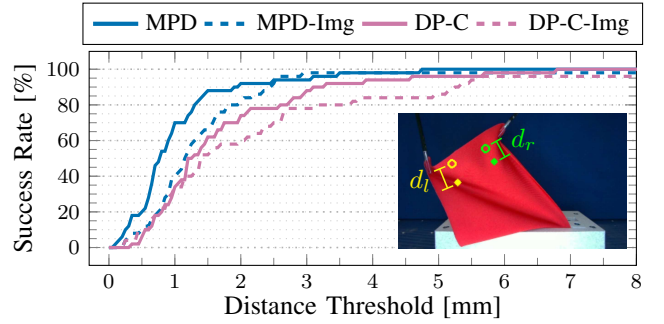

    Consistent with the motion metric results described below, DP-T and BESO generate considerably jagged action sequences.
    \gls{mpd} and DP-C generate overall smooth motion, however, DP-C often shows some jaggedness at the end of the task.
    On the simulated \gls{btm} task, success is defined by a fixed accuracy of aligning the markers and targets.
    However, in surgical scenarios the desired accuracy is often situational and changes across interventions and surgical phases~\cite{furuse2023influence, clarke2010measuring}.
    Thus, for the real world \gls{btm} task, success rate is not reported as a single value, but determined in relation to the desired accuracy for aligning target and marker points.
    The results for success rate in relation to desired accuracy are shown in \autoref{fig:success_rate_over_threshold}.
    The trajectory is successful if the distances $d_l$ and $d_r$ between left and right markers and targets are both below a threshold.

    \gls{mpd} outperforms DP-C for both image and state observations up to \SI{6.5}{\mm} accuracy, after which both reach $100\,\%$ success rate.
    For \SI{2}{\mm} accuracy and state observations, \gls{mpd} reaches $90\,\%$ while DP-C reaches $70\,\%$ success rate.
    The success of an episode on the real world \gls{rt}-RW, \gls{ll}-RW, and \gls{glt}-RW tasks is evaluated manually.
    Both \gls{mpd} and DP-C reach $100\,\%$ success rate on the real world tasks, with the exception of a success rate of $96\,\%$ for \gls{mpd} on \gls{rt}-RW.
    
\begin{figure}
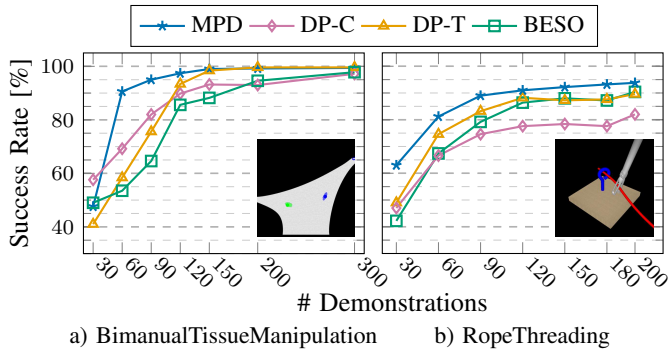

\centering
\vspace{1.5mm}
\begin{tikzpicture}
\begin{axis}[
    name=BTMNumTraj,
    width=0.6\columnwidth,
    height=0.48\columnwidth,
    xlabel={\# Demonstrations},
    xlabel style={
        at={(1.1, -0.15)},
    },
    ylabel={Success Rate [\%]},
    ylabel style={
        at={(-0.15, 0.5)},
    },
    ytick={0,20,...,100},
    xtick={30,60,90,120,150,200,300},
    xmin=20,
    xmax=310,
    ymin=30,
    ymax=105,
    legend style={
        at={(1.0,1.03),
        font=\small,
    },
    anchor=south,
    legend columns=4},
    ymajorgrids=true,
    yminorgrids=true,
    minor ytick={0,5,...,100},
    major grid style={semithick, black!30!white, dashed},
    minor grid style={ultra thin, black!20!white, dashed},
    minor y tick style = transparent,
    x tick label style={
        font=\footnotesize,
        rotate=-45,
        anchor=west,
    },
    y tick label style={
        font=\footnotesize,
    },
]

\addplot[color=MPD, mark=star, thick] coordinates {
    (30, 47.6)
    (60, 90.6)
    (90, 95.0)
    (120, 97.4)
    (150, 99.0)
    (200, 99.2)
    (300, 99.4)
};
\addlegendentry{MPD}

\addplot[color=DPC, mark=diamond, thick] coordinates {
    (30, 57.6)
    (60, 69.2)
    (90, 82.0)
    (120, 89.8)
    (150, 93.2)
    (200, 93.0)
    (300, 97.2)
};
\addlegendentry{DP-C}

\addplot[color=DPT, mark=triangle, thick] coordinates {
    (30, 41.0)
    (60, 58.4)
    (90, 75.6)
    (120, 93.4)
    (150, 98.4)
    (200, 99.6)
    (300, 99.6)
};
\addlegendentry{DP-T}

\addplot[color=BESO, mark=square, thick] coordinates {
    (30, 49.0)
    (60, 53.5)
    (90, 64.6)
    (120, 85.6)
    (150, 88.2)
    (200, 94.6)
    (300, 97.8)
};
\addlegendentry{BESO}

\end{axis}
\begin{axis}[
    name=RTNumTraj,
    at=(BTMNumTraj.south east),
    anchor=left of south west,
    width=0.6\columnwidth,
    height=0.48\columnwidth,
    ylabel style={
        at={(-0.08, 0.5)},
    },
    ytick={0,20,...,100},
    minor ytick={0,5,...,100},
    yticklabels={,,},
    xtick={30,60,90,120,150,180,200},
    xmin=20,
    xmax=220,
    ymin=30,
    ymax=105,
    ymajorgrids=true,
    yminorgrids=true,
    major grid style={semithick, black!30!white, dashed},
    minor grid style={ultra thin, black!20!white, dashed},
    minor y tick style = transparent,
    x tick label style={
        font=\footnotesize,
        rotate=-45,
        anchor=west,
    },
]

\addplot[color=BESO, mark=square, thick] coordinates {
    (30, 42.2)
    (60, 67.4)
    (90, 79.2)
    (120, 86.4)
    (150, 88.0)
    (180, 87.2)
    (200, 90.4)
};

\addplot[color=DPT, mark=triangle, thick] coordinates {
    (30, 49.0)
    (60, 74.6)
    (90, 83.2)
    (120, 88.2)
    (150, 87.25)
    (180, 87.6)
    (200, 89.6)
};

\addplot[color=DPC, mark=diamond, thick] coordinates {
    (30, 47.0)
    (60, 66.6)
    (90, 74.6)
    (120, 77.6)
    (150, 78.4)
    (180, 77.6)
    (200, 82.0)
};

\addplot[color=MPD, mark=star, thick] coordinates {
    (30, 63.0)
    (60, 81.2)
    (90, 89.0)
    (120, 91.0)
    (150, 92.2)
    (180, 93.2)
    (200, 93.8)
};
\end{axis}

\node[anchor=north] at ($(BTMNumTraj.south) + (0, -0.85cm)$) {\small a) BimanualTissueManipulation};
\node[anchor=north] at ($(RTNumTraj.south) + (0, -0.85cm)$) {\small b) RopeThreading};

\node[anchor=south east] at ($(BTMNumTraj.south east) + (0, 0.1cm)$) {\includegraphics[width=1.3cm]{images/tasks/BTM_end.eps}};
\node[anchor=south east] at ($(RTNumTraj.south east) + (0, 0.1cm)$) {\includegraphics[width=1.3cm]{images/tasks/RT_end.eps}};

\end{tikzpicture}

\caption{
    Success rates of MPD and the baselines over the number of training demonstrations on the Bimanual Tissue Manipulation and Rope Threading tasks on state observations.
}
\label{fig:num_traj}
\end{figure}

    \textbf{Data Efficiency:}
    All methods are trained multiple times using varying numbers of demonstrations.
    After capture, the demonstrations are shuffled once to mitigate the impact of possible learning curves in the demonstrations.
    For the experiments, the first $x$ demonstrations are used for training.
    The results for \gls{btm} and \gls{rt} tasks with state observations are shown in \autoref{fig:num_traj}.
    Compared to the baselines, \gls{mpd} requires much fewer demonstrations to reach high success rates.
    There is no clear trend as to which baseline method is more data efficient.
    For \gls{btm}, \gls{mpd} achieves around $90\,\%$ success rate for $90$ demonstrations.
    The other methods require $120$ for DP-T and $200$ for DP-C and BESO to reach similar success rates.

    \textbf{Motion Metrics:}
    We evaluate the methods as described in \autoref{sec:methods:hyper} on the motion metrics defined in \autoref{sec:experiments:metrics} on image observations and present the results in \autoref{fig:metrics}.

    \input{plots/motion_metrics}
    The values are normalized by human demonstration data, so values below $1.0$ indicate better-than-demonstrator motion quality.
    The transformer-based methods, namely BESO and DP-T, show high values for all motion quality metrics, except for episode length.
    Compared to BESO, DP-T differs noticeably across different runs, with min and max values that deviate far from the mean, see \eg, tissue acceleration in \autoref{fig:metrics} b).
    We report min and max values for the motion metrics instead of the standard deviation to highlight best- and worst-case scenarios for the methods.
    The results across different runs are more bounded for \gls{mpd} and DP-C, with min and max values closer to the mean.
    DP-C and especially \gls{mpd} achieve gentle motions whose values are much closer to the human demonstrations on all metrics.
    \gls{mpd} performs best across all metrics except for episode length on \gls{rt}.
    Using \glspl{prodmp}, \gls{mpd} generates motions that have noticeably less instrument jerk, even compared to human demonstrations.
    
\section{DISCUSSION}
\label{sec:discussion}
    High success rates of image-based policies are crucial for advancing the level of automation in \gls{ras}, because image observations are readily available in real surgical settings and do not require complex, task-specific feature extraction to generate state observations~\cite{scheikl2023sim2real}.
    However, we evaluated \gls{mpd} on both image and state observations to show that the benefits of our method are not limited to image observations.
    All investigated imitation learning methods reach much higher success rates on the same tasks, compared to policies trained with \gls{rl}, reported in~\cite{scheikl2023lapgym}.
    However, within the baseline methods there seems to be a trade-off between execution speed, success rate, and motion quality.
    DP-T, as a transformer-based \gls{ddpm} baseline reaches the highest baseline success rates, but generates motions that are unfavorable under surgical quality metrics.
    DP-C employs a 1D temporal CNN model architecture that generates smoother motions compared to the transformer-based baselines.
    However, the method does not explicitly prevent jagged motions and is noticeably slower during inference.
    Training speed is mainly influenced by the model architecture.
    For $3\,000$ epochs, the transformer-based methods require $400$ minutes while DP-C requires $200$ minutes on average for the real world datasets on a RTX 4090 GPU.
    In our experiments, training usually converges at around $400$ epochs, enabling training in less than an hour.
    \gls{mpd} and BESO benefit from the \gls{sgm} diffusion framework and can be executed with \SI{90}{\Hz}.
    With the same architecture, DP-T reaches \SI{80}{\Hz} while DP-C reaches \SI{40}{\Hz} due to a computationally more complex architecture. 
    The \gls{sgm} framework is therefore favorable for robotic applications that require online motion generation.
    However, BESO reaches the lowest average success rate on the evaluated tasks.
    In contrast to the baselines, \gls{mpd} features the fast inference time of the \gls{sgm} framework, the high task success rate of transformer-based methods, and improves on the motion quality of temporal CNN models for gentle manipulation of deformable objects.
    The gentle motions of \gls{mpd} are due to the use of \glspl{prodmp} but have an advantage over other \gls{il} methods that also utilize \glspl{prodmp}~\cite{li2023prodmp} in versatility, as \gls{mpd} is able to learn multimodal behaviors.

    The main limitation of \gls{mpd} lies in the limited performance outside of the available demonstration data, a common challenge for \gls{il} methods, that may be addressed by continued training with feedback~\cite{chi2020collaborative}.
    This limitation is apparent in the two failed episodes of the \gls{rt}-RW task, where initial threading of the rope failed and bent the rope so severely that the policy could not recover from this state.
    While the tasks are visually simple, surgically more realistic scenes are not a limitation as the relevant visual features are learned end-to-end.

    In preliminary experiments, we observed that the transformer-based methods are considerably more sensitive to their respective hyper parameter choices compared to DP-C.
    These findings are consistent with the findings of~\cite{chi2023diffusionpolicy}.
    Compared to BESO and DP-T, \gls{mpd} was less sensitive to the hyper parameters chosen.
    Furthermore, we observed that predicting more action steps than are executed, as described in~\cite{chi2023diffusionpolicy}, did not perform well on the tasks in this work.

    \gls{mpd} further outperforms all baselines in data efficiency, requiring fewer demonstrations to reach high success rates.
    The baseline methods exhibit a slight decrease in performance when increasing demonstrations from $150$ to $180$ on the \gls{rt} task, as shown in \autoref{fig:num_traj} b).
    This result may indicate that, compared to the baselines, \gls{mpd} performs better when added demonstrations introduce additional modes into the training data or that \gls{mpd} performs better on suboptimal demonstrations.
    Future work may investigate the behavior of \gls{mpd} under these two aspects.

\section{CONCLUSION}
\label{sec:conclusion}
    This work proposes \gls{mpd}, a novel method for learning robotic manipulation of deformable objects in the context of \gls{ras}.
    \gls{mpd} combines the versatility of \gls{dil} with the motion quality of \glspl{prodmp}, facilitating gentle manipulation of deformable objects and data efficient training that are crucial for surgical applications.
    The experiments show the superior performance of \gls{mpd} over traditional \gls{dil} methods.
    \gls{mpd} achieves higher success rates and further requires fewer demonstrations to reach comparable success rates.
    \Gls{mpd} shows favorable motion qualities, thereby aligning closely with the requirements for gentle tissue manipulation in surgical settings.
    The integration of \glspl{prodmp} allows generating smooth, high-frequency action sequences with guaranteed initial conditions, without an increase in inference time, required for application in real world robotic scenarios.

    In summary, \gls{mpd} offers a data efficient, gentle, and robust method for the manipulation of deformable objects in the context of \gls{ras}.
    Its ability to learn from limited data while maintaining high-quality motion characteristics makes it a promising approach for future developments in autonomous and semi-autonomous surgical systems.
    Future research may explore the application of \gls{mpd} in more diverse surgical scenarios and its integration with other surgical technologies, further pushing the boundaries of robotic assistance in complex medical procedures.

\bibliographystyle{IEEEtran.bst}
\bibliography{references}

\end{document}